\documentclass[lettersize,journal]{IEEEtran}
\usepackage{amsmath,amsfonts}
\usepackage{algorithmic}
\usepackage{algorithm}
\usepackage{array}
\usepackage[caption=false,font=normalsize,labelfont=sf,textfont=sf]{subfig}
\usepackage{textcomp}
\usepackage{stfloats}
\usepackage{url}
\usepackage{verbatim}
\usepackage{graphicx}
\usepackage{cite}
\usepackage{multirow}
\usepackage{booktabs}
\usepackage{color}
\usepackage{orcidlink}\hypersetup{colorlinks=true,linkcolor=black}
\usepackage{amsmath}

\begin{document}

\title{Multi-view feature High-order Fusion for Space Weak Object Detection and Segmentation}

\author{Weilong Guo$^*$~\orcidlink{0000-0002-9259-1746}, Yuhan Sun$^*$~\orcidlink{0000-0002-2493-2166} and Shengyang Li~\orcidlink{0000-0002-9888-9869} 
\thanks{This work was funded by the “Space Science and Application of China Manned Space Engineering DataBase” in the National Basic Science Data Center, Grant NO.NBSDC-DB-17. (\textit{Corresponding author: Shengyang Li. $^*$: Equal contribution})}
\thanks{Weilong Guo, Yuhan Sun and Shengyang Li are with the Technology and Engineering Center for Space Utilization, Chinese Academy of Sciences, Beijing 100094, China 
and Key Laboratory of Space Utilization, Chinese Academy of Sciences, Beijing 100094, China. 
Shengyang Li and Yuhan Sun are also with the University of Chinese Academy of Sciences, Beijing 100049, China (e-mail: guoweilong19@mails.ucas.ac.cn;  sunyuhan21@csu.ac.cn;shyli@csu.ac.cn).}
}

\markboth{Journal of \LaTeX\ Class Files,~Vol.~14, No.~8, August~2021}%
{Shell \MakeLowercase{\textit{et al.}}: A Sample Article Using IEEEtran.cls for IEEE Journals}


\maketitle

\begin{abstract}
Weak objects are common in images and videos of space applications. However, it is hard to learn proper representations from their limited appearance information. Inspired by multi-view learning, we develop simple multi-view attentions, treating their outputs as multi-view features. We also propose a multi-view feature high-order fusion method (MHF) to aggregate more accurate and richer features of weak objects. Our MHF extends the commonly used low-order feature fusion method to higher orders. It enhances the model's capacity to capture relevant and complementary information about weak objects. This is achieved by introducing high-order multi-view features perception and a recursive task-contribution gated selection of multi-view features. The new operation is highly flexible and customizable. It is compatible with various variants of multi-view feature representations. We conduct extensive experiments on two newly constructed space science datasets and an open, large-scale satellite video dataset. Our MHF serves as a plug-and-play module and significantly improves various vision transformers and convolution-based detection and segmentation models. We achieve all state-of-the-art accuracies on both tasks across three datasets. Our MHF can be a new basic module for visual modeling that effectively represents weak objects in terms of multi-view learning. The code will be available at https://github.com/Kingdroper/MHF.

\end{abstract}


\section{Introduction}
\IEEEPARstart{B}{enefiting} from the rapid development of multi-media techniques, the same object/image can usually be represented in multiple ways~\cite{ref2}. For example, the process of a scientific experiment in a space station can be represented using optical images, infrared images, density images, etc.~\cite{yangJ24}; a person can be identified by his/her face, fingerprints, and palmprint images~\cite{ref3,ref4}, and a single image can also be described by different types of features. In general, these data are referred to as multi-view data~\cite{ref5,ref6,ref7,ref8}, which has attracted significant attention in a wide range of applications~\cite{ref9}, e.g., surveillance~\cite{ref10}, autonomous driving~\cite{ref11}, and scientific research. In contrast to conventional single-view-based methods, multi-view learning enjoys the capability of extracting correlation and complementary information, contributing to performance improvement in many fields~\cite{ref2}.

\begin{figure*}[!htbp]
    \centering
    \includegraphics[width=5.5in]{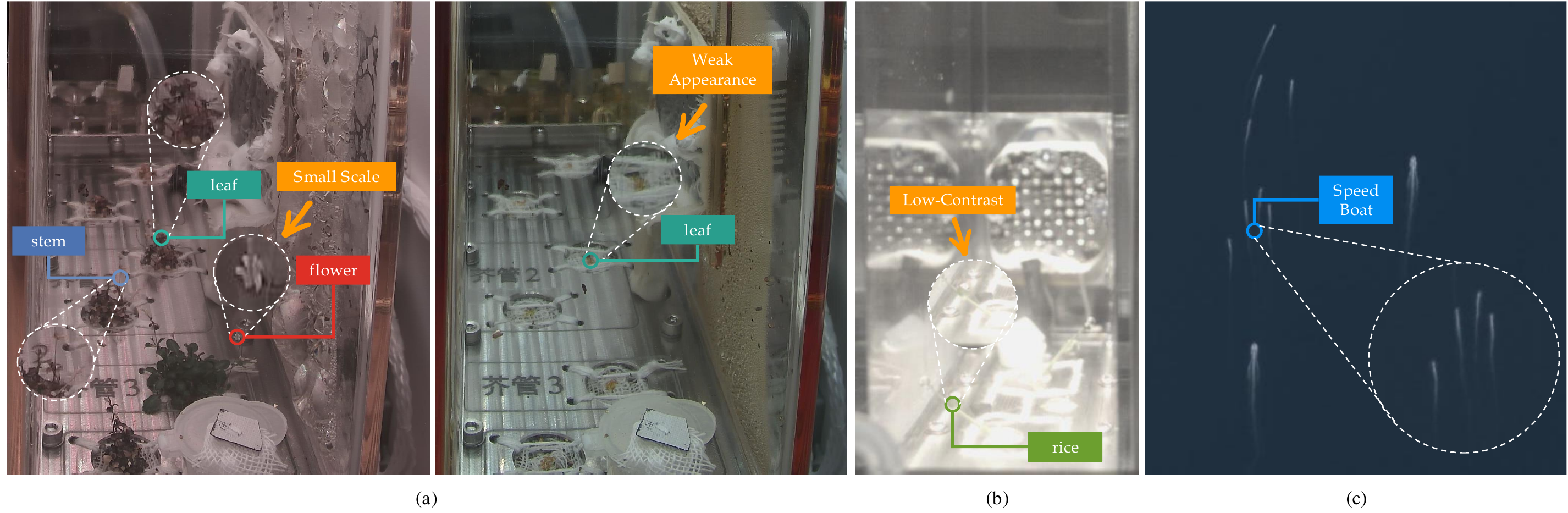}
    \caption{Examples of weak objects in space applications. (a) and (b) are images of plant (Arabidopsis and rice) growth experiments in the Chinese space station and TianGong2 space laboratory. (c) is a video of the Jilin-03 satellite.}
    \label{fig1}
\end{figure*}

Limited by the space environment, weak objects are common in many space science and earth observation applications. Some examples are shown in Figure~\ref{fig1}. Acquiring aligned multi-view data can be challenging. In this study, we concentrate on fusing the multi-view features of a single-modal image, with the image being acquired under a single-view condition. We consider different features of the same image as multi-view information and aim to design a multi-view feature fusion method to obtain more accurate and richer features of weak objects and to improve the detection and segmentation accuracy.

Some methods use different prior features, such as HoG, SIFT, and LBP as a multi-view representation~\cite{ref12,ref13}, but they are heavily influenced by prior knowledge and are not applicable to weak objects. We note that attention is a widely used feature representation mechanism in computer vision with high generality. The same attention design is applicable to processing data across different domains and can be easily integrated into various downstream models \cite{ref14}. It has already delivered strong performance in multiple fields, including earth observation \cite{ref15,ref16,ref17}, bioinformatics \cite{ref18}, medical science \cite{ref19,ref20}, and smart cities \cite{ref21,ref22}.

Existing attention can be broadly classified into spatial attention, channel attention, temporal attention, frequency attention, and transformer-like attention~\cite{ref23}. They are biased toward feature representations from different perspectives, respectively. For example, spatial attention and channel attention are biased towards contextual and semantic information, but successive convolution and pooling tend to lead to the loss of high-frequency details~\cite{ref24}. Frequency attention, however, effectively preserves high-frequency details, especially edge and boundary information~\cite{ref25,ref26}. Different types of attention can be seen as features of distinct perspectives on the same object. Some studies have demonstrated that combining different attentions can achieve better performance~\cite{ref27,ref28,ref29,ref30,ref31}. Most of the existing methods develop various simple fusion strategies~\cite{ref32}, such as the weighted average~\cite{ref33,ref34,ref35}, summation~\cite{ref36,ref37}, and concatenation~\cite{ref38,ref39}. Straightforward feature fusion lacks the explicit fusion principle to preserve the typical view feature~\cite{ref32}. Methods based on summation and concatenation are prone to feature redundancy, leading to model learning bias on redundant features and weak robustness. There is still a lack of a general and effective fusion strategy exploring the collaboration between multi-view attentions to further improve the accuracy of weak object detection and segmentation.

Although many fusion strategies have been explored in tasks such as multi-view learning~\cite{ref42} and multi-modality image fusion~\cite{ref40,ref41}. But they either require multi-modality data or did not have a fusion strategy that is detailed enough for weak objects. Weak object detection and segmentation require accurate, relevant, and complementary information selection and fusion to the target task. 

In this paper, we propose a plug-and-play module named multi-view feature high-order fusion (MHF) for weak object detection and segmentation in space applications. Specifically, we develop simple and effective spatial attention, channel attention, and frequency attention, and treat their outputs as multi-view features. Our MHF captures relevant and complementary information about weak objects by introducing high-order multi-view features perception and recursive task-contribution gated selection of multi-view features (CGS-n).  To avoid feature redundancy in the fusion stage, it learns the contribution of multi-view features to the detection and segmentation task in the normalized space and selects the task-relevant and complementary information among them. It also increases the correlation between the learned features and the detection and segmentation task. Theoretically, other types of multi-view features or combinations of different attention mechanisms can be fused using our MHF in a higher-order manner. The new operation offers high flexibility and customizability, and it remains compatible with various variants of multi-view feature representations. 

To evaluate our MHF, we constructed the M3SSE-ara and TianGong2-rice space science datasets using data from the Chinese Space Station and the TianGong2 space laboratory, respectively. Both of them include object detection and segmentation tasks. Small-scale, low-contrast, weak-looking objects are common. We perform extensive experiments on them and an open, large-scale satellite video dataset. Our MHF serves as a plug-and-play module and significantly improves 21 vision transformers and convolution-based detection and segmentation models. Without considering the cost of multi-view attentions, our MHF only costs less than 2M additional parameters. All state-of-the-art accuracies are achieved on both tasks of the three datasets. The high-order fusion manner makes our MHF easily adaptable to different application scenarios, and we only need to fine-tune the order according to the data distribution. These results clearly demonstrate the effectiveness and generality of our designs. Our MHF can be a new basic module for visual modeling that effectively represents weak objects in terms of multi-view learning.

Our main contributions are summarized as:
\begin{enumerate}\setcounter{enumi}{0}
\item We propose a new plug-and-play multi-view features high-order fusion method to aggregate more accurate and rich features of weak objects, extending the commonly used one-order/two-order feature fusion manner to high-order.

\item We design a recursive task-contribution gated selection of multi-view features, which can effectively avoid the common feature redundancy problem in feature fusion. It learns the contribution of multi-view features to weak object detection and segmentation in the normalized space and selects task-relevant and complementary information from coarse to fine in multiple steps.

\item We constructed two space science datasets using data from the Chinese Space Station and the TianGong2 space laboratory, conducting extensive experiments on them and an open large-scale satellite video dataset where weak objects are common. Our MHF serves as a plug-and-play module and significantly improves 21 vision transformers and convolution-based detection and segmentation models without bells and whistles.
\end{enumerate}

The methods most similar to our MHF are high-order attention (HA)~\cite{ref43} and HorNet~\cite{ref44}. For the input $X$, HA obtains Q, K, V, and an attention map in a self-attention way. Then it uses a threshold to force the values in the attention map to 0 and 1 as $B_1$. And the n power of $B_1$ is $B_n$. It is multiplied by the V to obtain features of different orders. Then they are fused by weighted addition. However, the features of different stages obtained in this way have high similarity and easily lead to feature redundancy. HorNet presents the $g^n$Conv that performs high-order spatial interactions with gated convolutions. It extends the two-order interactions in self-attention to arbitrary orders without introducing significant extra computation. It utilizes convolutions to lightly implement high-order spatial interactions like transformers.

Our MHF has the capability of high-order multi-view feature interaction, and we mainly focus on the high-order selection and fusion of multi-view features. We aim to fuse the information that is relevant and complementary to the weak object detection and segmentation task from the multi-view features to improve the performance in space applications. To the best of our knowledge, our MHF is the first lightweight method that enables arbitrarily high-order multi-view feature fusion.

\section{Related Works}
In space applications, most models are derived from generic detection and segmentation methods. In this section, we first review the general detection and segmentation methods, then summarize the commonly used strategies in weak object detection and segmentation, and finally summarize the methods of attention.

\subsection{General object detection and segmentation}
Object detection and segmentation aim to obtain the accurate locations and category information of objects of interest from images and videos. This information for object detection is object-level, while that for segmentation is pixel-level. The pipelines of existing methods can be uniformly divided into one-stage and two-stage. The one-stage methods usually directly generate the object-level or pixel-level information of objects, such as RetinaNet~\cite{ref45}, FCOS~\cite{ref46}, RepPoints~\cite{ref47}, and YOLACT~\cite{ref48}, etc. The two-stage methods first generate candidate regions where the objects may exist, and then predict accurate object-level or pixel-level information based on the candidates, such as RCNN families~\cite{ref49,ref50,ref51,ref52,ref53,ref54,ref55,ref56}, Hybrid Task Cascade (HTC)~\cite{ref57}, PointRend~\cite{ref58}, etc. Normally, two-stage methods have higher detection and segmentation accuracies, but their inference speeds are slower. Benefiting from the development of transformers recently, some transformer-based detection and segmentation methods, such as DETR~\cite{ref59}, DINO~\cite{ref60}, and Mask DINO~\cite{ref61}, etc. have emerged, reporting SOTA results on some mainstream benchmarks.

\subsection{Weak object detection and segmentation}
Weak objects are usually small-scale, low-contrast, or have a distorted appearance. It is hard to learn a discriminative representation from the limited information about them. Existing methods to address these problems include resolution-improvement-based methods and context-modeling-based methods.

The resolution-improvement-based methods aim at restoring the distorted structures of weak objects. Commonly used methods include deconvolutions, super-pixel convolutions~\cite{ref62}, self-supervision learning paradigms~\cite{ref63}, GANs~\cite{ref64,ref65,ref66,ref67}, etc. These methods reconstruct high-resolution weak objects to obtain more abundant and detailed information to improve object detection and segmentation accuracy.

The context-modeling-based methods are dedicated to exploring and integrating contextual cues of weak objects to model discriminative features. The most used contexts include multi-scale features from different layers of a CNN model, convolution features of different kernels~\cite{ref68,ref69}, features of dilated convolutions~\cite{ref70}, and features of different pooling~\cite{ref71}, etc. The current researches mainly focus on issues like the selection of context regions, encoding of context features, and the relationship between weak objects and contexts. However, there is no unified consensus on how to select local or global contexts for weak objects, and it is still in a heuristic and empirical fashion~\cite{ref72}.

\subsection{Attention mechanism}
Methods for diverting attention to the most important regions of an image and disregarding irrelevant parts are called attention mechanisms~\cite{ref23}. In a vision system, an attention mechanism can be treated as a dynamic selection process that is realized by adaptively weighting features according to the importance of the input. The popular attention currently available mainly includes channel attention and spatial attention~\cite{ref14,ref23}.

In deep neural networks, different channels in different features usually represent different objects~\cite{ref73}. Channel attention adaptively recalibrates the weight of each channel and can be viewed as an object selection process~\cite{ref23}. Hu et al.~\cite{ref74} first proposed the concept of channel attention and presented SENet for this purpose. It plays the role of emphasizing important channels while suppressing noise. SE blocks have shortcomings in the squeeze and excitation modules. Later channel attentions attempt to improve the outputs of the squeeze module~\cite{ref75}, reduce the complexity~\cite{ref76}, or improve both the squeeze module and the excitation module~\cite{ref77}. 

Spatial attention can be seen as an adaptive spatial region selection mechanism. RAM~\cite{ref78}, STN~\cite {ref79}, GENet~\cite{ref80}, and Non-local~\cite{ref81} are representative of different kinds of spatial attention methods. RAM represents RNN-based methods. STN represents those who use a subnetwork to explicitly predict relevant regions. GENet represents those that use a subnetwork implicitly to predict a soft mask to select important regions. Non-local represents self-attention-related methods. 

Most existing work focuses on the design of attention structures. Although some existing works try to combine the advantages of channel attention and spatial attention, their fusion is mostly one- or two-order, neglecting the exploitation of high-order information. Our MHF treats different attention as multi-view features and focuses on the high-order fusion of them to obtain better accuracy on weak object detection and segmentation.

\section{Method}
\begin{figure*}[!htbp]
    \centering
    \includegraphics[width=6.2in]{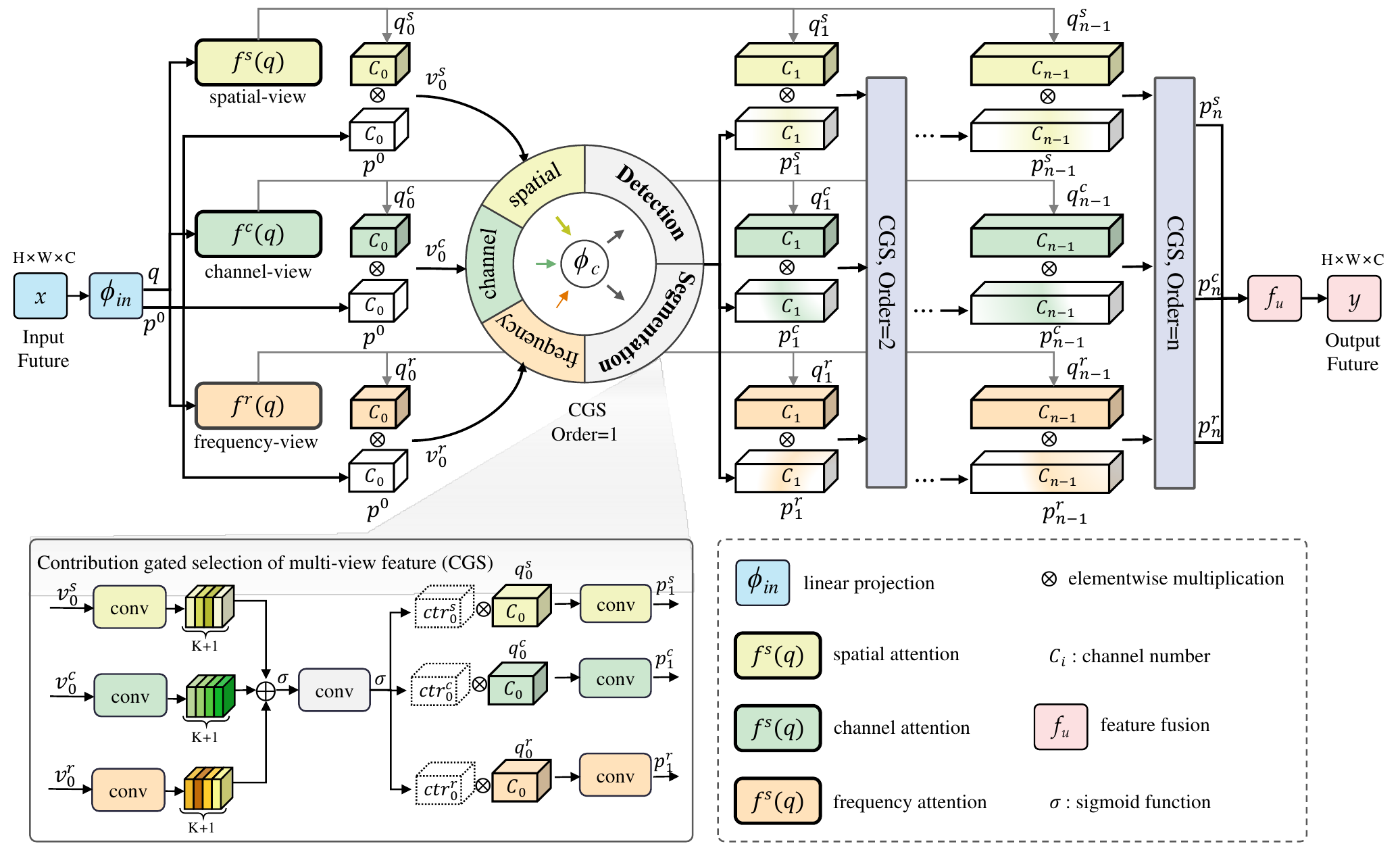}
    \caption{The detailed structure of our MHF.}
    \label{fig2}
\end{figure*}
In this section, we will present MHF, a lightweight plug-and-play module to achieve high-order fusion of multi-view attention features in a coarse-to-fine manner. Our MHF is built with standard convolutions, linear projections, and element-wise multiplications, but has a function of multi-view feature adaptive mixing. It outputs the feature $y$ that has the same size as the input feature $x$. As a result, it can be inserted between any two convolution layers. The overall pipeline is shown in Figure~\ref {fig2}.

\subsection{Multi-view attentions.}
Attention has been an important research field in the last few years, thriving in many different machine learning and vision applications~\cite{ref14}, such as object detection~\cite{ref16,ref17}, segmentation~\cite{ref25}, and so on. 

In this section, we develop simple and effective spatial attention, channel attention, and frequency attention, treating their outputs as multi-view features. Theoretically, we can use any of the existing attentions as multi-view features. All we need to do is just ensure that the attention used can represent the features of the image from different perspectives. Since designing new attention is not the focus of this paper, we only use simple attention representations, and some designs refer to existing methods.
\begin{figure}[!htbp]
    \centering
    \includegraphics[width=3.3in]{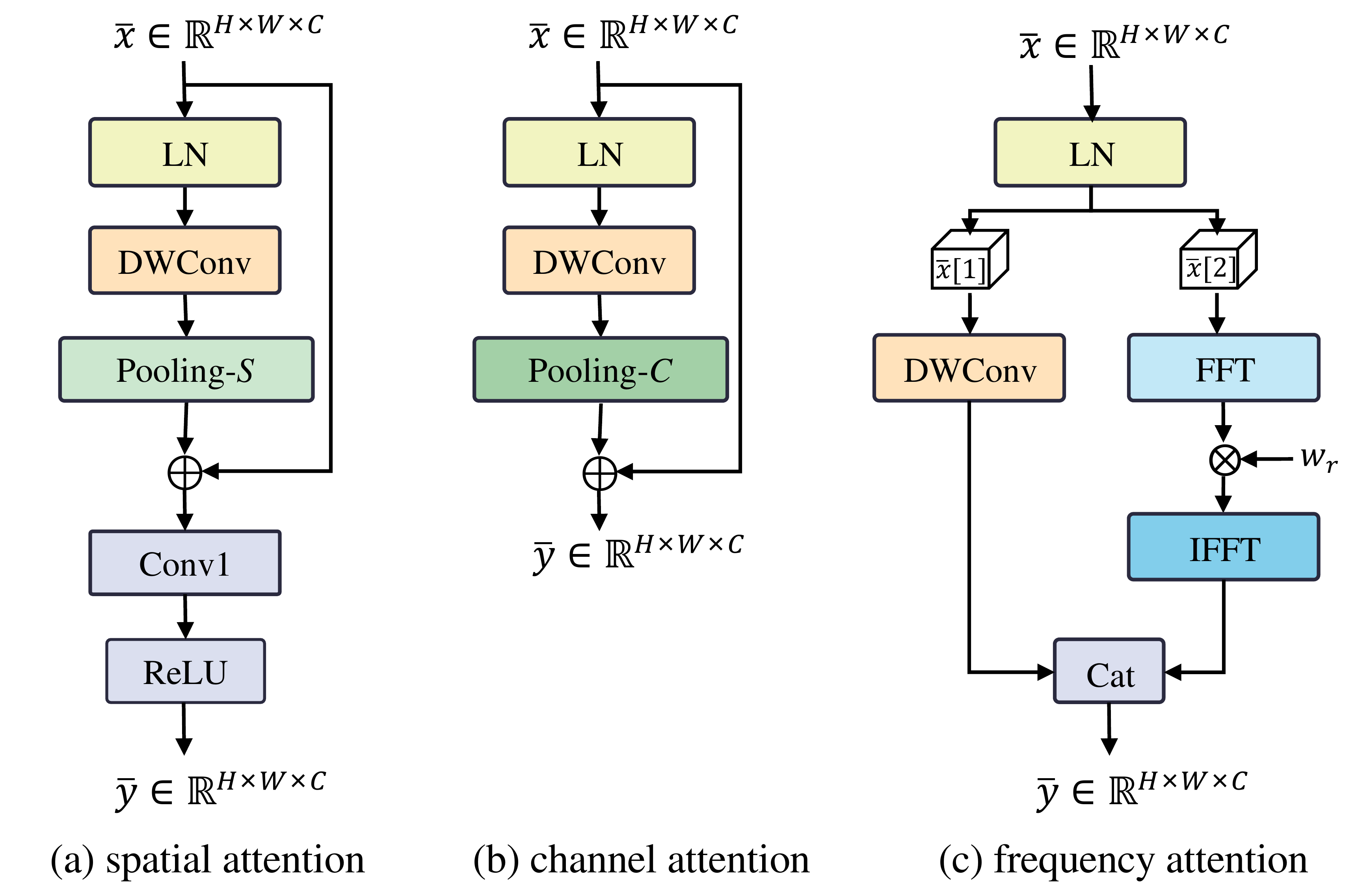}
    \caption{The detailed structure of our multi-view attentions. LN is the layer normalization. DWConv is a depth-wise convolution layer and the kernel size is $3\times 3$. Pooling-$S$ and Pooling-$C$ are average pooling of spatial and channel dimensions. Conv1 is a $1\times 1$ convolution layer. ReLU is an activation function. Cat denotes the operation of stacking features along the channel dimension. FFT and IFFT mean the fast Fourier transform and inverse fast Fourier transform, respectively.}
    \label{fig3}
\end{figure}

The detailed structures of our multi-view attention are shown in Figure~\ref {fig3}. For different attention, our design follows different principles. For spatial attention and channel attention, the kernel idea is to endow each region of the input feature with global and learnable information in spatial and channel dimensions, respectively. The input feature $\overline{x}$ is first regularized by layer normalization. Then we use a depth-wise convolution layer to assign learnable weights to each position of the feature while capturing local information. The kernel size is $3\times 3$. The main difference between our spatial and channel attention is the capture of global information. For their lightness, we use the average pooling of spatial and channel dimensions as the global information, respectively.  The results are added to the input feature $\overline{x}$ through the residual connection, respectively. This gives them a computational complexity of $O(n)$. Most of the background information in the image is distracting to weak objects, so in our spatial attention, we further adaptively suppress it using a $1\times 1$ convolution with a ReLU activation function. The semantic information of weak objects is very limited, and the information represented by each channel may be useful for subsequent weak object detection and segmentation. Unlike the existing channel attention, we do not further select the channels. 

For frequency attention, representing image information in the frequency domain is as important as aligning features. Similarly, we first regularise the input feature $\overline{x}$ using layer normalization. We then divide the regularised features equally along the channel dimension into features $\overline{x}[1]$ and $\overline{x}[2]$. The $\overline{x}[1]$ is fed into the depth-wise convolution for obtaining local context with a kernel size of $3\times 3$. A 2D fast Fourier transform (FFT) is performed on $\overline{x}[2]$ to convert it to the frequency domain. And a global learnable weight $w_r$ is used to learn frequency-view features who has the same size as $\overline{x}[2]$. The $w_r$ is essentially a learnable complex tensor. Interaction and fusion with features from spatial and channel attention are required in subsequent multi-view feature selection and fusion. The features from frequency attention should be spatially aligned. Therefore, the output is converted back to the spatial domain by inverse fast Fourier transform (IFFT). The outputs of the two branches are stacked along the channel dimension to obtain spatially aligned frequency features.

\subsection{Contribution gated selection of multi-view feature (CGS)}
Our CGS aims to select relevant and complementary information of weak objects from multi-view features by sensing their contribution to the target task. It mainly includes the perception and selection of multi-view features. The recent success of visual transformers and transformer-like attentions relies heavily on the proper modeling of spatial interactions in visual data. However, the $O(n^2)$ complexity of the kernel multi-head self-attention hinders application in downstream tasks, especially object detection and segmentation where high-resolution feature maps are required. Recent work like HorNet~\cite{ref44} demonstrates that global spatial interactions can be efficiently implemented using CNNs. Inspired by this, we extend such spatial interactions to arbitrary multi-view features. While different from HorNet, our CGS can collaboratively achieve multi-view feature perception and selection according to the target task.

Let $x \in \mathbb{R}^{HW\times C}$ be the input feature, the output of multi-view feature perception can be written as:
\begin{gather}
    \label{eq1}
    [p_0\in \mathbb{R}^{HW\times C}, q_0\in \mathbb{R}^{HW\times C}] = \phi _{in}(x)\in \mathbb{R}^{HW\times C} \\
    v_0^* = f^*(q_0)\otimes p_0\in \mathbb{R}^{HW\times C}
\end{gather}
where $\phi _{in}$ is the linear projection layer to perform channel mixing and $f^*$ is our multi-view attentions. $*\in \{s, c, r\}$. $f^s$, $f^c$, and $f^r$ are spatial attention, channel attention, and frequency attention, respectively. $v_0^*$ is the multi-view perception feature. $v_0^{*,(i,c)}=\sum _{j\in \Omega_i}f^*(q_0)^{(j,c)}p_0^{(i,c)}$. $\Omega _i$ is the perception window at $i$ for multi-view attention features. The above equation shows the interaction between $p_0^i$ and the multi-view attention feature $f^* (q_0)^j$ through element-wise multiplication $\otimes $.  $i$ and $j$ denote the location index of the feature and $c$ denotes the channel index of the feature. Each position of our multi-view attention encodes a global-local context. Then the feature interaction can adaptively capture the information related to $p_0$ in the multi-view features. Our CGS then learns the contribution of $v_0^*$ to the weak object detection and segmentation task, and based on this selects complementary information of weak objects from multi-view features.
\begin{gather}
    \label{eq2}
    {ctr}_0^s,{ctr}_0^c,{ctr}_0^r = \phi _c (v_0^s,v_0^c,v_0^r) \\
    p_1^*=\phi _{out}({ctr}_0^*\otimes f^*(q_0)\in \mathbb{R}^{HW\times C}
\end{gather}

$\phi _c$ is the task contribution learning network. It is built with convolutions, activation functions, and element-wise multiplications. Object detection and segmentation aim to determine the probability that each feature value belongs to the different objects, $(K+1)$. $K$ denotes the number of categories, and 1 refers to the background. As shown in Figure~\ref{fig2}, our CGS first projects $v_0^*$ to this probability space separately through convolution layers with the sigmoid activation function. Then it collects and stacks the results along the channels and feeds them to a convolution with an activation function to generate the contributions ${ctr}_0^*$ to weak object detection and segmentation. Finally, it performs multi-view feature selection by element-wise multiplications with $f^*(q_0)$. $\phi _{out}$ is the linear projection layer to perform channel mixing. We consider the selection of multi-view feature in CGS as 1-order as $p_0$ has interacted with $f^* (q_0)$ once and performed a feature selection based on the contribution to the weak object detection and segmentation task.

\subsection{High-order contribution gated selection of multi-view feature (CGS-n)}
After achieving 1-order multi-view feature selection CGS, we then design CGS-n, a recursive task-contribution gated multi-view feature selection module, which is the kernel of our MHF. It further enhances the model's capacity to capture relevant and complementary information about weak objects by introducing high-order multi-view features perception and recursive task-contribution gated selection of multi-view features. Formally, We use $\phi_{in}$ to obtain projection features $p_0\in \mathbb{R}^{HW\times C_0}$ and $q\in \mathbb{R}^{HW\times \sum_{0\leq k\leq n-1} C_k }$. We take $q$ as input to obtain multi-view attention features and split them into a set of features $\{q_k^*\}_{k=0}^{n-1}$ along the channel dimension. $*\in \{s,c,r\}$ denotes the spatial view, channel view, and frequency view, respectively. 
\begin{equation}
\begin{aligned}
    \label{eq3}
    &[q_0^*\in \mathbb{R}^{HW\times C_0},\cdots, q_{n-1}^*\in \mathbb{R}^{HW\times C_{n-1}}] \\
    &= f^*(q)\in \mathbb{R}^{HW\times \sum_{0\leq k\leq n-1} C_k}
\end{aligned}
\end{equation}
We then perform multi-view feature perception and selection recursively by
\begin{gather}
    \label{eq4}
    v_k^* = q_k^*\otimes p_k^*\in \mathbb{R}^{HW\times C_k} \\
    \label{eq4_1}
    {ctr}_k^s,{ctr}_k^c,{ctr}_k^r = \phi _c (v_k^s,v_k^c,v_k^r) \\
    \label{eq4_2}
    p_{k+1}^*=\phi _{out}^k({ctr}_k^*\otimes q_k^*)\in \mathbb{R}^{HW\times C_{k+1}}\\
    \label{eq4_3}
    \phi_{out}^k=\begin{cases}
        None, & k=0 \\
        Linear(C_k,C_{k+1}), & 1\leq k\leq {n-1}
    \end{cases}
\end{gather}

$\phi_{out}^k$ is a linear projection layer to perform channel mixing and match the dimensions in different orders. From the above recursive formula equations, it is easy to show that the multi-view feature perception and selection order will be increased by 1 after each step. As a result, we can see that the CGS-n achieves n-order multi-view feature perception and selection with contribution gating. It is also worth noting that we need only a single $f^*$ to perform multi-view attention and a single $\phi_c$ to perform task contribution sensing instead of computing/building them in each recursive step as in equation~\ref{eq4}-\ref{eq4_2}, which can further simplify the implementation and improve the efficiency on GPUs. To ensure that the high-order multi-view feature perception and selection do not introduce too much computational overhead, we set the channel dimension in each order as:
\begin{gather}
    \label{eq5}
    C_k=\frac{C}{2^{n-k-1}}, 0\leq k\leq {n-1}
\end{gather}

Besides, the channel dimension of $\phi_{in} (x)$ is exactly $2C$, and the total FLOPs can be strictly bounded even with $n$ increasing. This design indicates that we perform the multi-view feature selection in a coarse-to-fine manner, where lower orders are computed with fewer channels.

\subsection{Feature Fusion}
After high-order multi-view feature perception and selection, we obtain weak object-related and complementary multi-view information $p_n^*$. Our CSG-n is designed to meticulously extract the complementary multi-view features of weak objects and effectively avoid feature redundancy. Therefore, no complex design is required in the final fusion part. We simply sum $p_{n}^*,*\in\{s,c,r\}$ and pass the result through a ReLU activation function as the final output.

\subsection{Plug-and-play}
Our MHF outputs the feature $y$ with the same size as the input feature $x$. As a result, it can be inserted between any two convolution layers. The structure of currently existing object detection and segmentation methods includes: (1) A feature extraction module including a backbone network such as ResNet~\cite{ref82}, DarkNet~\cite{ref83}, and a multi-scale feature utilization network (neck) such as FPN~\cite{ref84}, and YOLONeck~\cite{ref85}. (2) Dense prediction heads such as classification head, regression head, and segmentation head. To validate the effectiveness of MHF, we insert it after the feature extraction module of current state-of-the-art object detection and segmentation methods and apply it to the output multi-scale features, respectively.

\section{Experiments}
In this section, we conduct extensive experiments to verify the effectiveness of our method. We present the main results on two constructed space science datasets, M3SSE-ara and TianGong2-rice, and an open large-scale satellite video dataset, SAT-MTB~\cite{ref86}, where weak objects are common, and make comparisons with various advanced object detection and segmentation methods. Lastly, we provide ablation studies of our designs and analyze the effectiveness of our MHF on a wide range of models.

\subsection{Datasets}
M3SSE-ara and TianGong2-rice are space science datasets. Scientists aim to explore the effects of microgravity environments on plant growth and place cameras on the space station to film Arabidopsis and rice growing. It is important for cultivating plants with new characteristics in the future. To help scientists analyze plant growth, we constructed the M3SSE-ara and TianGong2-rice datasets using data from the Chinese Space Station and the TianGong2 space laboratory, respectively. Both of them include object detection and segmentation tasks, and small-scale, low-contrast, weak-looking objects are common. The filming period of M3SSE-ara is from July 2022 to November 2022. It requires detecting and segmenting the flowers, stems, and leaves of the Arabidopsis in each image. It includes a total of 3720 images, where 2980 images are used for training and 740 images are used for testing. The image sizes are all $1300\times 1160$ after preprocessing. TianGong2-rice is a rice leaf detection and segmentation dataset. It requires detecting and segmenting each leaf of the rice in the images. Its images are captured during the rice growth experiments from September 2016 to March 2017. It includes a total of 800 images, where 640 images are used for training and 160 images are used for testing. The image sizes are all $1000\times 400$. Its challenge is mainly that the contrast between the leaves of the rice and the background is so low that it is difficult to detect and segment them. In the future, these two datasets will be made publicly available in our Multi-field, Multi-modal, and Multi-task benchmark for Space Science Experiments (M3SSE) open project\footnote{\label{m3s}http://www.csu.cas.cn/gb/kybm/sjlyzx/gcxx\_sjj/sjj\_tgxl}.

To further validate the effectiveness of our method, we perform experiments on the open-source large-scale satellite video dataset, SAT-MTB~\cite{ref86}, as well. It is currently the largest multi-task benchmark for satellite video, including object detection, segmentation, and tracking. A large amount of the objects are small-scale, faint-appearing objects. Its videos are from the Jilin-03 satellite. We verify the effectiveness of our MHF on the object detection and segmentation tasks of the dataset. It includes a total of 144 videos, 33228 frame images, and 308204 objects. It requires detecting and segmenting objects of 12 fine-grained classes including Freighter (FH), Rear engine aircraft (RA), Train (TN), Naval vessels (NV), Yacht (YC), Narrow-body aircraft (NA), Cruise (CS), Four engine aircraft (FA), Wide body aircraft (WA), Speed boat (SB), Corporate aircraft (CA), and Other ship (OS). The frame image size of SAT-MTB ranges from $500\times 600$ to $1500\times 3000$.

\subsection{Evaluation Metrics and Comparison Methods}
We use the mainstream COCO evaluation metrics~\cite{ref87} to evaluate object detection and segmentation performance of different methods on M3SSE-ara and TianGong2-rice datasets, including mAP, $\rm mAP_{50}$, $\rm mAP_{75}$, $\rm AP_s$, $\rm AP_m$, and $\rm AP_l$. The mAP is the mean of average precision under different IoU thresholds, 0.5:0.95:0.05. The $\rm mAP_{50}$ and $\rm mAP_{75}$ are the means of average precision when the IoU thresholds are 0.5 and 0.75, respectively. The $\rm AP_s$, $\rm AP_m$, and $\rm AP_l$ metrics are used to evaluate the detection and segmentation accuracy of small, medium, and large objects, respectively. In this paper, we focus on mAP, $\rm mAP_{50}$, and $\rm AP_s$ associated with weak objects. On the SAT-MTB dataset, we follow the benchmark settings~\cite{ref86} and use $\rm mAP_{50}$ to evaluate the overall detection and segmentation accuracy of different methods. The $\rm AP_{50}$ is used to evaluate the fine-grained detection and segmentation accuracy of each category.

We mainly compare with the current advanced object detection and segmentation methods on the above three datasets. It includes one- and two-stage methods~\cite{ref45,ref46,ref49,ref50}, transformer-based methods~\cite{ref59,ref60,ref61}, and video-based methods. Their codes come from the open-source mmdetection\footnote{\label{mdet}https://github.com/open-mmlab/mmdetection} and mmtracking\footnote{\label{mtk}https://github.com/open-mmlab/mmtracking} 
 framework. 

\begin{table*}[!thbp] 
    \scriptsize
    \renewcommand\arraystretch{1.1}
    \centering
    \caption{Comparisons with state-of-the-art object detection and segmentation methods on the M3SSE-ara dataset. \label{tab1}}
    \setlength{\tabcolsep}{0.95mm}{
    \begin{tabular}{c|c|c|c|c|cccccc|ccc}
    
    \specialrule{0.1em}{1pt}{1pt}
    \multirow{2}{*}{\textbf{Method}} & {\textbf{Accepted}} & {\textbf{Feature}} & {\textbf{Epoch/}} & {\textbf{MHF}} & \multirow{2}{*}{$\rm mAP$} & \multirow{2}{*}{$\rm mAP_{50}$} & \multirow{2}{*}{$\rm mAP_{75}$} & \multirow{2}{*}{$\rm AP_s$} & \multirow{2}{*}{$\rm AP_m$} & \multirow{2}{*}{$\rm AP_l$} & \multicolumn{3}{c}{AP of Classes}\\
    \cline{12-14}
    {} & {\textbf{by}} & {\textbf{Extraction}} & {\textbf{steps}} & {\textbf{(ours)}} & {} & {} & {} & {} & {} & {} & {leaf} & {stem} & {flower} \\
    \hline
    \multicolumn{14}{c}{\textbf{Object Detection}} \\
    \hline
    {Faster RCNN~\cite{ref49}} & {TPAMI2017} & {ResNet101+FPN} & {84} & {} & {66.8} & {95.9} & {78.2} & {64.4} & {75.8} & {75} & {69.9} & {69.3} & {61.1} \\
    \hline
    {CornerNet~\cite{ref88}} & {ECCV2018} & {HourglassNet104} & {210} & {} & {36.4} & {59.4} & {38.6} & {37.4} & {51.8} & {42.8} & {34} & {33.6} & {41.5} \\
    \hline
    {FCOS~\cite{ref46}} & {ICCV2019} & {ResNet101+FPN} & {84} & {} & {52.7} & {88.5} & {59.7} & {47.8} & {68.8} & {62.8} & {56.7} & {48.3} & {53} \\
    \hline
    {CenterNet~\cite{ref89}} & {ArXiv2019} & {ResNet18+FPN} & {84} & {} & {11.1} & {30.7} & {6.1} & {6.1} & {25.4} & {37.1} & {16.3} & {8} & {9} \\
    \hline
    \multirow{2}{*}{Empirical Attn.(0010)~\cite{ref90}} & \multirow{2}{*}{ICCV2019} & {ResNet50+FPN} & {84} & {} & {64} & {95.4} & {74.8} & {61.5} & {71.6} & {74.8} & {66.5} & {64.2} & {61.2} \\
    {} & {} & {ResNet50+FPN+DCN} & {84} & {} & {64.3} & {95.4} & {75.8} & {61.5} & {72.7} & {74.9} & {68.1 } & {63.2 } & {61.5} \\
    \hline
    {Empirical Attn.(1111)~\cite{ref90}} & {ICCV2019} & {ResNet50+FPN} & {84} & {} & {65.0 } & {95.4} & {77.2} & {62.6} & {72.4} & {76.6} & {68.3} & {64.9} & {61.7} \\
    \hline
    {CentripetalNet~\cite{ref91}} & {CVPR2020} & {HourglassNet104} & {210} & {} & {61.3} & {91.9} & {70.7} & {54.6} & {79.2} & {84.6} & {66} & {57.1} & {60.9} \\
    \hline
    \multirow{2}{*}{YOLOv3~\cite{ref83}} & \multirow{2}{*}{ICCV2018} & \multirow{2}{*}{DarkNet53} & {273} & {} & {63} & {94.3} & {72} & {58.9} & {73.8} & {74} & {66.3} & {62.6} & {60.2} \\
    {} & {} & {} & {273} & {\checkmark} & {65.8} & {94.7} & {76.7} & {61.8} & {77.3} & {76.3} & {67.7} & {65.7} & {64.1} \\
    \hline
    \multirow{2}{*}{FSAF~\cite{ref92}} & \multirow{2}{*}{CVPR2019} & \multirow{2}{*}{ResNet101+FPN} & {84} & {} & {62.8} & {89.0 } & {73.2} & {57.3} & {75.7} & {78.3} & {68.0 } & {56.7} & {63.7} \\
    {} & {} & {} & {84} & {\checkmark} & {68.8} & {90.3} & {77.1} & {64.2} & {81.1} & {84.0 } & {72.6} & {65.2} & {68.6} \\
    \hline
    \multirow{2}{*}{ATSS~\cite{ref93}} & \multirow{2}{*}{CVPR2020} & \multirow{2}{*}{ResNet101+FPN} & {84} & {} & {60.3} & {87.6} & {68.3} & {52.9} & {80.3} & {78.9} & {63.4} & {57.2} & {60.5} \\
    {} & {} & {} & {84} & {\checkmark} & {67.8} & {88.5} & {75.2} & {61.4} & {85.1} & {82.5} & {71.0 } & {66.6} & {65.9} \\
    \hline
    \multirow{2}{*}{Dynamic RCNN~\cite{ref55}} & \multirow{2}{*}{ECCV2020} & \multirow{2}{*}{ResNet101+FPN} & {84} & {} & {62.0 } & {88.9} & {71.7} & {59.0 } & {71.4} & {76.8} & {65} & {64.3} & {56.7} \\
    {} & {} & {} & {84} & {\checkmark} & {71.8} & {93.8} & {83.2} & {69.3} & {79.7} & {78.5} & {72.5} & {73.7} & {69.3} \\
    \hline
    \multirow{2}{*}{FoveaBox~\cite{ref94}} & \multirow{2}{*}{TIP2020} & \multirow{2}{*}{ResNet101+FPN} & {84} & {} & {52} & {71.8} & {58.4} & {43.5} & {78.2} & {76.0 } & {56.6} & {48.9} & {50.5} \\
    {} & {} & {} & {84} & {\checkmark} & {54.4} & {70.8} & {60.8} & {45.7} & {81.4} & {82.3} & {58.4} & {53.5} & {51.3} \\
    \hline
    \multirow{2}{*}{Cascade RCNN~\cite{ref50}} & \multirow{2}{*}{TPAMI2021} & \multirow{2}{*}{ResNet101+FPN} & {84} & {} & {51} & {65.9} & {57.7} & {40.9} & {82.8} & {83.2} & {54.1} & {49.9} & {49.2} \\
    {} & {} & {} & {84} & {\checkmark} & {52} & {64.5} & {58.2} & {42.2} & {82.9} & {83.3} & {56.6} & {52.9} & {46.6} \\
    \hline
    \multirow{2}{*}{DINO~\cite{ref60}} & \multirow{2}{*}{ICLR2023} & \multirow{2}{*}{ResNet101+FPN} & {84} & {} & {74.8} & {95.2} & {84.1} & {72.2} & {83.7} & {83.9} & {74.9} & {74.1} & {70.4} \\
    {} & {} & {} & {84} & {\checkmark} & {76.3} & {95.0} & {84.8} & {74.2} & {85.6} & {83.8} & {77.1} & {81.4} & {70.8} \\
    \hline
    \multicolumn{14}{c}{\textbf{Segmentation}} \\
    \hline
    {HTC~\cite{ref57}} & {CVPR2019} & {ResNet101+FPN} & {84} & {} & {33.7} & {50.9} & {37.3} & {25.3} & {59.5} & {43.0} & {31.3} & {27.1} & {42.8} \\
    \hline
    {YOLACT~\cite{ref48}} & {ICCV2019} & {ResNet101+FPN} & {84} & {} & {14.5} & {37.5} & {8} & {9.1} & {33.5} & {38.8} & {20.2} & {4.3} & {19.1} \\
    \hline
    {SCNet~\cite{ref95}} & {AAAI2021} & {ResNet101+FPN} & {84} & {} & {41.8} & {53.2} & {47.3} & {31.9} & {69.4} & {76.5} & {33.9} & {43.7} & {47.7} \\
    \hline
    \multirow{2}{*}{Mask RCNN~\cite{ref51}} & \multirow{2}{*}{ICCV2017} & \multirow{2}{*}{ResNet101+FPN} & {84} & {} & {53.7} & {90.0} & {57.2} & {52.3} & {60.3} & {42.0 } & {61.2} & {40.9} & {59.1} \\
    {} & {} & {} & {84} & {\checkmark} & {57.6} & {91.1} & {63.0 } & {57.1} & {64.6} & {44.4} & {65.4} & {47.4} & {60.1} \\
    \hline
    \multirow{2}{*}{MaskScoring RCNN~\cite{ref96}} & \multirow{2}{*}{CVPR2019} & \multirow{2}{*}{ResNet101+FPN} & {84} & {} & {55.5} & {90.2} & {61} & {55.1} & {60.6} & {44.1} & {61.4} & {44.7} & {60.3} \\
    {} & {} & {} & {84} & {\checkmark} & {62.5} & {91.6} & {70.1} & {62.6} & {67.4} & {45.1} & {68.5} & {53.3} & {65.7} \\
    \hline
    \multirow{2}{*}{PointRend~\cite{ref58}} & \multirow{2}{*}{CVPR2020} & \multirow{2}{*}{ResNet101+FPN} & {84} & {} & {56.6} & {92.2} & {61.4} & {54.8} & {64.1} & {48.2} & {61.4} & {45.1} & {63.3} \\
    {} & {} & {} & {84} & {\checkmark} & {59.3} & {92.3} & {65.6} & {57.9} & {67.9} & {48.9} & {64.7} & {49.4} & {63.8} \\
    \hline
    \multirow{2}{*}{QueryInst~\cite{ref97}} & \multirow{2}{*}{ICCV2021} & \multirow{2}{*}{ResNet101+FPN} & {84} & {} & {43.9} & {78.3} & {45.7} & {38.6} & {60.4} & {49.7} & {45.8} & {40.5} & {45.3} \\
    {} & {} & {} & {84} & {\checkmark} & {62.3} & {93.2} & {70.0 } & {60.1} & {71.2} & {57.4} & {67.2} & {60.0 } & {59.8} \\
    \hline
    \multirow{2}{*}{Cascade Mask RCNN~\cite{ref50}} & \multirow{2}{*}{TPAMI2021} & \multirow{2}{*}{ResNet101+FPN} & {84} & {} & {36.0 } & {62.2} & {36.2} & {28.4} & {62.5} & {43.6} & {44.5} & {23.7} & {39.9} \\
    {} & {} & {} & {84} & {\checkmark} & {41.4} & {62.1} & {45.5} & {33.5} & {71.5} & {45.4} & {51.4} & {29.9} & {43.0} \\
    \hline
    \multirow{2}{*}{SparseInst~\cite{ref98}} & \multirow{2}{*}{ICLR2023} & \multirow{2}{*}{ResNet101+FPN} & {27K} & {} & {34.4} & {65.7} & {33.6} & {30.6} & {52.0} & {42.5} & {49.1} & {0.8} & {46.2} \\
    {} & {} & {} & {27K} & {\checkmark} & {35.1} & {64.9} & {35.4} & {31.1} & {56.4} & {42.1} & {49.5} & {6.6} & {49.1} \\
    \specialrule{0.1em}{1pt}{1pt}
    \end{tabular}}
    \end{table*}

\subsection{Implement Details and Setups}
All comparison experiments and ablation studies are conducted on an Nvidia Tesla V100 GPU, whose memory is 32 GB. The code of our MHF is implemented on the PyTorch framework.

On the M3SSE-ara and TianGong2-rice datasets, to preserve as much information about the weak objects as possible, we use the original size of the image as input. And we initialize all methods using the pre-trained models on the COCO dataset~\cite{ref87}. Method settings, such as learning rate, data enhancement process, etc. follow the default settings of mmdetection if not specified. On the M3SSE-ara dataset, we train 84 epochs for methods using the SGD optimizer and use ResNet101 to extract features for most methods using ResNet as the backbone network. For methods using the Adam or AdamW optimizers, and methods using other backbone networks, we follow their default settings. On the TianGong2-rice dataset, since most of the objects are small-scale and low-contrast, we train 120 epochs for all methods and use ResNet50+FPN to extract multi-scale features. On the SAT-MTB dataset, we follow the settings on its object detection and segmentation benchmarks~\cite{ref86}. Except for the latest DINO detection method, which is trained for 270k steps, we train all other detection methods for 20 epochs, and segmentation methods for 12 epochs. In each step, we randomly crop 512 × 512 regions from input images for training. When different methods are equipped with our MHF, it is trained together with other parameters of the model. Our MHF has a hyperparameter, order. When our MHF is applied to different methods, the order is uniformly set to 2 on the M3SSE-ara and TianGong2-Rice datasets, and 6 on the SAT-MTB dataset. In the ablation studies, we further explore the impact of different orders on the accuracy of different methods and datasets.

\begin{table*}[!thbp] 
    \footnotesize
    \renewcommand\arraystretch{1.0}
    \centering
    \caption{Comparisons with state-of-the-art object detection and segmentation methods on the TianGong2-rice dataset.\label{tab2}}
    \setlength{\tabcolsep}{0.9mm}{
    \begin{tabular}{c|c|c|c|cccccc}
        
    \specialrule{0.1em}{1pt}{1pt}
    \multirow{2}{*}{\textbf{Method}} & {\textbf{Accepted}} & {\textbf{Feature}} & {\textbf{MHF}} & \multirow{2}{*}{$\rm mAP$} & \multirow{2}{*}{$\rm mAP_{50}$} & \multirow{2}{*}{$\rm mAP_{75}$} & \multirow{2}{*}{$\rm mAP_s$} & \multirow{2}{*}{$\rm mAP_m$} & \multirow{2}{*}{$\rm mAP_l$} \\
    {} & {\textbf{By}} & {\textbf{Extraction}} & {\textbf{(ours)}} & {} & {} & {} & {} & {} & {} \\
    \hline
    \multicolumn{10}{c}{\textbf{Object Detection}} \\
    \hline
    {Dynamic RCNN~\cite{ref55}} & {ECCV2020} & \multirow{22}{*}{ResNet50+FPN} & {} & {52.1} & {80.4} & {55.6} & {59.3} & {55.5} & {4.8}  \\
    {GFL~\cite{ref99}} & {NeurIPS2020} & {} & {} & {28.5} & {61.4} & {25.3} & {30.8} & {41.7} & {4.6}  \\
    {DyHead~\cite{ref100}} & {CVPR2021} & {} & {} & {3.0} & {11.3} & {0.6} & {7.1} & {22.1} & {0.1}  \\
    {DAB-DETR~\cite{ref101}} & {ICLR2022} & {} & {} & {12.5} & {38.4} & {6.1} & {13.5} & {20.5} & {0.1}  \\
    \cline{1-2} \cline{4-10} 
    \multirow{2}{*}{Faster RCNN~\cite{ref49}} & \multirow{2}{*}{TPAMI2017} & {} & {} & {49.1} & {77.6} & {53.4} & {56.4} & {51.5} & {6.8}  \\
    {} & {} & {} & {\checkmark} & {51.8} & {77.6} & {56.4} & {61.7} & {54.4} & {9.6}  \\ 
    \cline{1-2} \cline{4-10}
    \multirow{2}{*}{FCOS~\cite{ref46}} & \multirow{2}{*}{ICCV2019} & {} & {} & {9.1} & {37.1} & {0.8} & {23.5} & {15.3} & {1.2}  \\
    {} & {} & {} & {\checkmark} & {18.0} & {53.3} & {6.4} & {33.1} & {23.4} & {0.5}  \\ 
    \cline{1-2} \cline{4-10}
    \multirow{2}{*}{FSAF~\cite{ref92}} & \multirow{2}{*}{CVPR2019} & {} & {} & {44.9} & {77.0} & {43.2} & {54.4} & {48.1} & {1.3}  \\
    {} & {} & {} & {\checkmark} & {53.1} & {80.3} & {55.2} & {61.6} & {56.4} & {1.9}  \\ 
    \cline{1-2} \cline{4-10}
    \multirow{2}{*}{DoubleHead~\cite{ref54}} & \multirow{2}{*}{CVPR2020} & {} & {} & {48.6} & {77.5} & {51.8} & {55.3} & {50.0} & {5.8}  \\
    {} & {} & {} & {\checkmark} & {51.9} & {75.0} & {59.4} & {62.9} & {53.0} & {9.0}  \\   
    \cline{1-2} \cline{4-10}
    \multirow{2}{*}{ATSS~\cite{ref93}} & \multirow{2}{*}{CVPR2020} & {} & {} & {54.9} & {82.1} & {59.5} & {63.9} & {58.4} & {6.7}  \\
    {} & {} & {} & {\checkmark} & {55.7} & {78.5} & {59.0} & {64.5} & {58.9} & {2.9}  \\ 
    \cline{1-2} \cline{4-10}
    \multirow{2}{*}{FoveaBox~\cite{ref94}} & \multirow{2}{*}{TIP2020} & {} & {} & {43.4} & {74.9} & {44.3} & {59.7} & {46.3} & {1.9}  \\
    {} & {} & {} & {\checkmark} & {47.8} & {74.8} & {50.5} & {60.6} & {51.7} & {6.4}  \\    
    \cline{1-2} \cline{4-10}
    \multirow{2}{*}{RetinaNet~\cite{ref45}} & \multirow{2}{*}{TPAMI2021} & {} & {} & {23.2} & {63.0} & {9.8} & {47.2} & {28.2} & {0.3}  \\
    {} & {} & {} & {\checkmark} & {41.5} & {80.1} & {35.9} & {55.1} & {47.9} & {2.0}  \\ 
    \cline{1-2} \cline{4-10} 
    \multirow{2}{*}{Libra RCNN~\cite{ref53}} & \multirow{2}{*}{IJCV2021} & {} & {} & {53.4} & {76.4} & {58.9} & {62.5} & {55.1} & {6.2}  \\
    {} & {} & {} & {\checkmark} & {55.4} & {79.7} & {60.3} & {63.3} & {56.5} & {11.1}  \\ 
    \cline{1-2} \cline{4-10}
    \multirow{2}{*}{DINO~\cite{ref60}} & \multirow{2}{*}{ICLR2023} & {} & {} & {57.3} & {80.0} & {62.2} & {64.7} & {63.7} & {10.0}  \\
    {} & {} & {} & {\checkmark} & {62.9} & {82.5} & {68.3} & {64.9} & {69.3} & {14.1}  \\   
    \hline
    \multicolumn{10}{c}{\textbf{Segmentation}} \\
    \hline
    {HTC~\cite{ref57}} & {CVPR2019} & \multirow{14}{*}{ResNet50+FPN} & {} & {6.9} & {17.7} & {3.1} & {15.1} & {5.3} & {6.7}  \\
    {Cascade Mask RCNN~\cite{ref50}} & {TPAMI2021} & {} & {} & {9.8} & {27.0} & {4.6} & {25.3} & {7.4} & {6.6}  \\
    {SCNet~\cite{ref95}} & {AAAI2021} & {} & {} & {9.3} & {22.2} & {4.7} & {19.8} & {6.9} & {3.4}  \\
    {DetectorRS~\cite{ref92}} & {CVPR2021} & {} & {} & {7.7} & {18.4} & {3.4} & {16.7} & {5.8} & {3.4}  \\
    \cline{1-2} \cline{4-10} 
    \multirow{2}{*}{Mask RCNN~\cite{ref51}} & \multirow{2}{*}{ICCV2017} & {} & {} & {16.3} & {42.3} & {7.7} & {32.0} & {13.1} & {23.4}  \\
    {} & {} & {} & {\checkmark} & {17.6} & {43.9} & {10.4} & {35.1} & {14.5} & {16.8}  \\ 
    \cline{1-2} \cline{4-10}
    \multirow{2}{*}{PointRend~\cite{ref58}} & \multirow{2}{*}{CVPR2020} & {} & {} & {14.3} & {41.9} & {4.8} & {30.2} & {11.3} & {20.1}  \\
    {} & {} & {} & {\checkmark} & {18.4} & {45.8} & {11.7} & {35.0} & {14.8} & {16.8}  \\ 
    \cline{1-2} \cline{4-10} 
    \multirow{2}{*}{MaskScoring RCNN~\cite{ref96}} & \multirow{2}{*}{IJCV2021} & {} & {} & {19.4} & {45.2} & {13.2} & {36.9} & {15.6} & {10.1}  \\
    {} & {} & {} & {\checkmark} & {23.3} & {51.1} & {18.0} & {39.9} & {19.8} & {13.5}  \\ 
    \cline{1-2} \cline{4-10} 
    \multirow{2}{*}{QueryInst~\cite{ref97}} & \multirow{2}{*}{ICCV2021} & {} & {} & {20.8} & {56.9} & {8.7} & {26.5} & {20.4} & {16.8}  \\
    {} & {} & {} & {\checkmark} & {25.6} & {62.6} & {12.6} & {27.5} & {25.7} & {20.1}  \\ 
    \cline{1-2} \cline{4-10} 
    \multirow{2}{*}{SparseInst~\cite{ref98}} & \multirow{2}{*}{ICLR2023} & {} & {} & {29.4} & {60.7} & {26.3} & {30.8} & {29.5} & {0.0}  \\
    {} & {} & {} & {\checkmark} & {32.0} & {64.5} & {31.0} & {32.1} & {32.3} & {10.1}  \\ 

    \specialrule{0.1em}{1pt}{1pt}
    \end{tabular}}
    \end{table*}

\subsection{Comparisons with State-of-the-art Methods}

\textbf{Results on M3SSE-ara and TianGong2-rice Dataset}. These two datasets include image sequences of Arabidopsis and rice growth taken inside the Chinese space station and the TianGong2 space laboratory. It reflects the appearance changes throughout the growth cycle. However, it is difficult to distinguish the parts of most plants, such as flowers, stems, and leaves, due to the effects of the space environment and growth stages, such as light. We use these two datasets on the one hand to validate the effectiveness of our method and on the other hand to promote new scientific discoveries in space science applications.

We adopt different advanced methods to perform object detection and segmentation using our MHF to evaluate its effectiveness. We insert our MHF after the feature extraction module of existing methods and apply it to the output multi-scale features, respectively. We use the original image as input without scaling and follow the default data pipeline and hyperparameters of different methods. All models are retrained on the training set and tested on the test set. The results on M3SSE-ara and TianGong2-rice datasets are summarized in table~\ref{tab1} and ~\ref{tab2} respectively. When our MHF is applied to various object detection and segmentation methods with different structures, including one-stage, two-stage, and transformer-based methods, their accuracies significantly improve under the same conditions. DINO with our MHF achieves the best detection accuracies, $76.3\%$ and $62.9\%$ mAP, on both datasets. Mask Scoring RCNN and SparseInst with our MHF achieves the best segmentation accuracies, $62.5\%$ and $32.0\%$ mAP, on two datasets, respectively. Our MHF also significantly improves the $\rm AP_s$ of various methods. These results clearly demonstrate the effectiveness and generality of our designs.

\textbf{Results on SAT-MTB Dataset}. To evaluate the effectiveness and robustness of our method in complex scenarios, we also test our models on the SAT-MTB dataset. It is the largest open satellite video multi-task benchmark. All model settings follow the benchmark. We apply our MHF to the top object detection and segmentation methods of the benchmark and report the overall accuracy,$\rm mAP_{50}$, and the accuracy, $\rm AP_{50}$, of each fine-grained category. The results are shown in Table~\ref {tab3}. 

 \begin{table*}[!thbp] 
        \footnotesize
        \renewcommand\arraystretch{1.0}
        \centering
        \caption{Comparisons with state-of-the-art object detection and segmentation methods on the SAT-MTB dataset.\label{tab3}}
        \setlength{\tabcolsep}{0.45mm}{
        \begin{tabular}{c|c|c|c|cccccccccccc|c}
        
        \specialrule{0.1em}{1pt}{1pt}
        \multirow{2}{*}{\textbf{Method}} & {\textbf{Accepted By}} & {\textbf{Feature}} & {\textbf{MHF}} & \multicolumn{12}{c|}{$\rm AP_{50}$ of Classes} & \multirow{2}{*}{$\rm mAP_{50}$} \\
        \cline{5-16}
        {} & {\textbf{By}} & {\textbf{Extraction}} & {\textbf{(ours)}} & {FH} & {RA} & {TN} & {NV} & {YC} & {NA} & {CS} & {FA} & {WA} & {SB} & {CA} & {OS} & {} \\
        \hline
        \multicolumn{17}{c}{\textbf{Object Detection}} \\
        \hline
        {FGFA~\cite{ref103}} & {ICCV2017} & \multirow{12}{*}{ResNet50+FPN} & {} & {12.6} & {69.3} & {34.5} & {2.2} & {5.1} & {42.1} & {1.3} & {42.0} & {68.7} & {0.0} & {0.7} & {15.9} & {24.5} \\
        {DFF~\cite{ref104}} & {CVPR2017} & {} & {} & {21.2} & {67.1} & {23.8} & {0.0} & {6.7} & {36.4} & {3.1} & {41.6} & {65.4} & {0.0} & {0.7} & {28.9} & {24.6} \\
        {SELSA~\cite{ref105}} & {ICCV2019} & {} & {} & {19.9} & {65.1} & {34.3} & {0.2} & {7.3} & {52.3} & {0.6} & {46.5} & {71.0} & {0.0} & {1.6} & {17.3} & {26.3} \\
        {Temporal RoI Align~\cite{ref106}} & {AAAI2021} & {} & {} & {6.3} & {44.3} & {22.6} & {0.1} & {3.4} & {35.4} & {4.6} & {33.4} & {55.3} & {0.1} & {0.0} & {3.1} & {17.4} \\
        \cline{1-2} \cline{4-17}
        {Grid RCNN~\cite{ref52}} & {CVPR2019} & {} & {} & {10.4} & {80.5} & {61.5} & {0} & {39.7} & {46.3} & {0} & {28.4} & {56.3} & {1.3} & {34.1} & {10.2} & {30.7} \\
        {TridentNet~\cite{ref107}} & {ICCV2019} & {} & {} & {0} & {66.8} & {52.3} & {0} & {19.1} & {54.4} & {0} & {44.6} & {61.2} & {2.6} & {33.4} & {0} & {27.9} \\
        {DETR~\cite{ref59}} & {ECCV2020} & {} & {} & {43.2} & {63.9} & {48.9} & {0.0} & {36.9} & {54.5} & {0.1} & {47.9} & {50.8} & {6.6} & {37.2} & {11.9} & {33.5} \\
        {Libra RCNN~\cite{ref53}} & {IJCV2021} & {} & {} & {32.7} & {47.9} & {52.4} & {0.2} & {33.6} & {43.0} & {0.0} & {29.9} & {47.9} & {0.1} & {13.8} & {8.9} & {25.8} \\
        {RetinaNet~\cite{ref45}} & {TPAMI2021} & {} & {} & {9.1} & {58.1} & {42.0} & {0.0} & {24.0} & {49.2} & {0.0} & {23.3} & {48.8} & {8.5} & {10.5} & {0.3} & {22.8} \\
        {Groie~\cite{ref108}} & {ICPR2021} & {} & {} & {3.3} & {72.8} & {59.2} & {0} & {42.6} & {50.7} & {0.1} & {36.9} & {59} & {8.1} & {24.6} & {13.8} & {30.9} \\
        \cline{1-2} \cline{4-17}
        \multirow{2}{*}{Faster RCNN~\cite{ref49}} & \multirow{2}{*}{TPAMI2017} & {} & {} & {9.8} & {64.9} & {42.8} & {0.0} & {32.1} & {62.0} & {0.0} & {36.6} & {52.2} & {0.7} & {20.8} & {2.2} & {27.0} \\
        {} & {} & {} & {\checkmark} & {19.0} & {74.3} & {54.9} & {0.0} & {50.9} & {53.5} & {0.0} & {39.6} & {60.3} & {17.2} & {29.8} & {11.4} & {34.2} \\
        \hline
        \multirow{2}{*}{YOLOv3~\cite{ref83}} & \multirow{2}{*}{ICCV2018} & \multirow{2}{*}{DarkNet53} & {} & {5.5} & {60.7} & {44.2} & {0.0} & {29.4} & {48.8} & {0.4} & {49.7} & {50.5} & {8.8} & {30.2} & {65.7} & {32.8} \\
        {} & {} & {} & {\checkmark} & {0.8} & {64.6} & {51.5} & {0.0} & {40.1} & {60.5} & {0.0} & {41.4} & {65.5} & {6.4} & {43.0} & {30.5} & {33.7} \\
        \hline
        \multirow{2}{*}{FCOS~\cite{ref46}} & \multirow{2}{*}{ICCV2019} & \multirow{8}{*}{ResNet50+FPN} & {} & {9.6} & {37.6} & {36.8} & {0.0} & {27.6} & {43.3} & {0.0} & {25.6} & {52.7} & {5.9} & {12.5} & {11.8} & {21.9} \\
        {} & {} & {} & {\checkmark} & {5.1} & {77.8} & {52.7} & {0.0} & {46.2} & {51.9} & {0.0} & {41.6} & {66.7} & {8.2} & {29.1} & {70.8} & {37.5} \\
        \cline{1-2} \cline{4-17}
        \multirow{2}{*}{ATSS~\cite{ref93}} & \multirow{2}{*}{CVPR2020} & {} & {} & {31.7} & {54.6} & {37.7} & {0.0} & {33.0} & {56.0} & {0.0} & {30.1} & {57.2} & {3.4} & {18.1} & {32.3} & {29.5} \\
        {} & {} & {} & {\checkmark} & {10.2} & {66.9} & {44.0} & {0.0} & {40.6} & {48.4} & {5.5} & {40.5} & {68.3} & {7.9} & {29.4} & {65.5} & {35.6} \\
        \cline{1-2} \cline{4-17}
        \multirow{2}{*}{Cascade RCNN~\cite{ref50}} & \multirow{2}{*}{TPAMI2021} & {} & {} & {24.8} & {61.6} & {55.1} & {0.0} & {28.3} & {58.7} & {0.0} & {26.4} & {51.5} & {0.4} & {21.7} & {1.5} & {27.5} \\
        {} & {} & {} & {\checkmark} & {18.7} & {65.8} & {51.5} & {0.0} & {41.7} & {57.2} & {0.0} & {38.8} & {66.6} & {2.1} & {44.5} & {75.2} & {38.5} \\
        \cline{1-2} \cline{4-17}
        \multirow{2}{*}{DINO~\cite{ref60}} & \multirow{2}{*}{ICLR2023} & {} & {} & {13.2} & {59.0} & {64.1} & {0.0} & {33.8} & {59.0} & {0.8} & {28.4} & {51.4} & {7.7} & {2.5} & {39.8} & {30.0} \\
        {} & {} & {} & {\checkmark} & {12.0} & {57.3} & {64.3} & {0.0} & {35.6} & {59.1} & {0.2} & {33.5} & {53.4} & {7.2} & {2.3} & {56.9} & {31.8} \\
        \hline
        \multicolumn{17}{c}{\textbf{Segmentation}} \\
        \hline
        {MaskScoring RCNN~\cite{ref96}} & {CVPR2019} & \multirow{4}{*}{ResNet50+FPN} & {} & {3.6} & {73.8} & {3.2} & {8.3} & {41.1} & {48.4} & {0.0} & {39.8} & {71.5} & {0.2} & {1.8} & {24.5} & {27.7} \\
        {YOLACT~\cite{ref48}} & {ICCV2019} & {} & {} & {0.0} & {19.5} & {0.0} & {0.0} & {26.4} & {11.3} & {0.0} & {37.4} & {42.6} & {0.6} & {4.9} & {21.0} & {13.6} \\
        {QueryInst~\cite{ref97}} & {ICCV2021} & {} & {} & {0.5} & {54.2} & {0.0} & {0.3} & {14.5} & {22.5} & {0.0} & {11.3} & {16.3} & {1.4} & {10.5} & {23.1} & {12.9} \\
        {DetectorRS~\cite{ref102}} & {CVPR2021} & {} & {} & {11.5} & {83.5} & {3.5} & {10.8} & {33.8} & {52.9} & {0.0} & {55.4} & {76.7} & {0.9} & {23.9} & {24.5} & {31.4} \\
        \hline
        {SeqFormer~\cite{ref109}} & {ECCV2022} & {\multirow{3}{*}{ResNet50}} & {} & {0.0} & {29.8} & {0.6} & {3.8} & {0.6} & {12.7} & {0.0} & {21.1} & {16.2} & {0.0} & {1.6} & {20.9} & {10.3} \\
        {IDOL~\cite{ref110}} & {ECCV2022} & {} & {} & {0.0} & {24.6} & {1.3} & {2.8} & {10.2} & {2.8} & {0.0} & {16.5} & {13.0} & {0.0} & {0.2} & {12.6} & {10.5} \\
        {Mask Transfiner~\cite{ref111}} & {CVPR2022} & {} & {} & {0.9} & {48.5} & {4.6} & {1.6} & {13.5} & {24.8} & {0.0} & {13.4} & {41.6} & {0.0} & {7.6} & {15.4} & {17.6} \\
        \hline
        {MTTR~\cite{ref112}} & {CVPR2022} & {Swin Transformer} & {} & {1.2} & {53.1} & {7.6} & {3.6} & {21.6} & {31.8} & {0.0} & {26.5} & {23.1} & {0.2} & {9.4} & {19.8} & {19.4} \\
        \hline
        \multirow{2}{*}{Mask RCNN~\cite{ref51}} & \multirow{2}{*}{ICCV2017} & {\multirow{10}{*}{ResNet50+FPN}} & {} & {4.8} & {60.5} & {1.6} & {15.5} & {36.9} & {52.7} & {0.0} & {40.8} & {66.9} & {0.0} & {24.8} & {14.1} & {26.6} \\
        {} & {} & {} & {\checkmark} & {0.0} & {64.5} & {3.7} & {36.8} & {45.2} & {56.3} & {0.0} & {31.9} & {61.3} & {4.1} & {45.9} & {24.2} & {31.2} \\
        \cline{1-2} \cline{4-17}
        \multirow{2}{*}{HTC~\cite{ref57}} & \multirow{2}{*}{CVPR2019} & {} & {} & {16.1} & {76.1} & {5.7} & {26.4} & {36.4} & {51.2} & {0.0} & {40.4} & {71.1} & {0.1} & {30.5} & {30.5} & {32.0} \\
        {} & {} & {} & {\checkmark} & {5.7} & {80.2} & {1.3} & {38.5} & {36.5} & {62.2} & {0.0} & {43.9} & {72.1} & {0.0} & {30.5} & {28.6} & {33.3} \\
        \cline{1-2} \cline{4-17}
        \multirow{2}{*}{PointRend~\cite{ref58}} & \multirow{2}{*}{CVPR2020} & {} & {} & {9.8} & {61.6} & {26.6} & {2.7} & {31.8} & {55.3} & {0.0} & {37.8} & {64.1} & {0.0} & {23.4} & {20.4} & {27.8} \\
        {} & {} & {} & {\checkmark} & {2.1} & {67.1} & {11.8} & {54.6} & {45.9} & {57.4} & {0.0} & {35.0} & {70.3} & {6.9} & {47.5} & {32.5} & {35.9} \\
        \cline{1-2} \cline{4-17}
        \multirow{2}{*}{Cascade Mask RCNN~\cite{ref50}} & \multirow{2}{*}{TPAMI2021} & {} & {} & {4.3} & {70.0} & {0.8} & {30.6} & {38.8} & {53.0} & {0.0} & {37.4} & {60.1} & {0.2} & {24.0} & {32.6} & {29.3} \\
        {} & {} & {} & {\checkmark} & {22.9} & {77.8} & {0.2} & {64.3} & {32.7} & {55.4} & {0.0} & {26.9} & {53.1} & {0.0} & {26.3} & {21.8} & {31.8} \\
        \cline{1-2} \cline{4-17}
        \multirow{2}{*}{SCNet~\cite{ref95}} & \multirow{2}{*}{AAAI2021} & {} & {} & {9.0} & {89.3} & {1.5} & {0.3} & {38.7} & {66.3} & {0.0} & {55.1} & {74.5} & {0.1} & {28.8} & {11.0} & {31.2} \\
        {} & {} & {} & {\checkmark} & {18.6} & {85.1} & {2.0} & {66.3} & {38.5} & {56.2} & {0.0} & {40.2} & {73.2} & {0.0} & {22.5} & {28.3} & {35.9} \\
        \specialrule{0.1em}{1pt}{1pt}
        \end{tabular}}
        \end{table*}

YOLOv3 uses the DarkNet53 as the backbone, MTTR uses the Swin Transformer [99] as the backbone, and the other methods use the ResNet50 as the backbone. For object detection, most image-based detection methods have a higher $\rm mAP_{50}$ than that of video-based methods such as Temporary RoI Align, SELSA, FGFA, DFF, etc. Most two-stage detection methods (e.g., Faster RCNN, Libra RCNN, and Cascade RCNN) have a higher $\rm mAP_{50}$ than that of one-stage methods (e.g., ATSS, FCOS, RetinaNet). The transformer-based method, DETR, achieves the highest detection accuracy of $\rm mAP_{50}$, which is $33.5\%$, among comparison methods. While the two-stage method, HTC, achieves the highest segmentation accuracy of $\rm mAP_{50}$, which is $32.0\%$.

When applying our MHF to different detection methods, including Faster RCNN, YOLOv3, ATSS, FCOS, and Cascade RCNN, their overall accuracy, $\rm mAP_{50}$, significantly improves. The improvement for FCOS is the most significant, and the $\rm mAP_{50}$ increases from $21.9\%$ to $37.5\%$ ($\uparrow 15.6\%$). The Cascade RCNN with our MHF achieves the highest $\rm mAP_{50}$ accuracy of $38.5\%$ among all detection methods. When our MHF is applied to different segmentation methods, including HTC, Mask RCNN, Cascade Mask RCNN, SCNet, and PointRend, the improvement of $\rm mAP_{50}$ accuracy is obvious. The PointRend method has the most visible improvement when equipped with our MHF, and the $\rm mAP_{50}$ accuracy increases from $27.8\%$ to $35.9\%$ ($\uparrow 8.1\%$). The PointRend and SCNet with our MHF conclude the highest $\rm mAP_{50}$ accuracy, which is $35.9\%$. These results show that our MHF significantly improves the fine-grained detection and segmentation accuracy of weak objects on satellite videos, where weak objects are common. It demonstrates the effectiveness and robustness of our method.

\subsection{Ablation Studies}
We provide detailed ablation studies of our MHF in table~\ref{tab4} and use the typical Mask RCNN as the baseline to study its designs. The backbone network is ResNet50. We first study the multi-view attentions and compare our MHF with existing fusion strategies on the M3SSE-ara dataset. Then, we explore the influences of different order settings of our MHF for different methods on different datasets. To further understand the superiority of our MHF in learning discriminative representations of weak objects, we compare it with the 11 latest and excellent attention models under fair conditions on the M3SSE-Ara dataset. These results clearly show that our MHF is an efficient and extendable operation that can better capture related complementary information of weak objects from multi-view features. 

\begin{table*}[!thbp] 
    \footnotesize
    \renewcommand\arraystretch{1.25}
    \centering
    \caption{Ablation study and results of equipping with different components on the M3SSE-ara dataset.
    The baseline method is Mask RCNN with the ResNet50 as the backbone.
    The training epochs are 84.
    $f^s$, $f^c$, and $f^r$ are our spatial attention, channel attention, and frequency attention respectively.
    The "Sum" means element-wise summation operation for feature fusion.
    The "Cat" indicates that the features are fused by stacking them along the channel dimension.
    \label{tab4}}
    \setlength{\tabcolsep}{0.9mm}{
    \begin{tabular}{ccc|ccc|cccccc|ccc}
        
    \specialrule{0.1em}{1pt}{1pt}
    \multirow{2}{*}{\textbf{$f^s$}} & \multirow{2}{*}{\textbf{$f^c$}} & \multirow{2}{*}{\textbf{$f^r$}} & \multicolumn{3}{c|}{Fusion Strategies} & \multirow{2}{*}{$\rm mAP$} & \multirow{2}{*}{$\rm mAP_{50}$} & \multirow{2}{*}{$\rm mAP_{75}$} & \multirow{2}{*}{$\rm AP_s$} & \multirow{2}{*}{$\rm AP_m$} & \multirow{2}{*}{$\rm AP_l$} & \multicolumn{3}{c}{AP of Classes}\\
    \cline{4-6} \cline{13-15}
    {} & {} & {} & {Sum} & {Cat} & {CGS-n(Ours)} & {} & {} & {} & {} & {} & {} & {leaf} & {stem} & {flower} \\
    \hline
    {} & {} & {} & {} & {} & {} & {51.6} & {89.8} & {52.0} & {49.6} & {58.8} & {42.1} & {58.3} & {38.3} & {58.0} \\
     {\checkmark} & {} & {} & {} & {} & {} & {54.3} & {90.7} & {58.4} & {53.3} & {59.3} & {42.3} & {61.8} & {40.4} & {60.7} \\
     {} & {\checkmark} & {} & {} & {} & {} & {54.7} & {90.5} & {59.3} & {53.3} & {61.5} & {43.6} & {62.2} & {40.6} & {61.4} \\
     {} & {} & {\checkmark} & {} & {} & {} & {55.0} & {89.7} & {60.1} & {54.1} & {61.0} & {42.9} & {62.4} & {42.1} & {60.5} \\
     {\checkmark} & {\checkmark} & {\checkmark} & {\checkmark} & {} & {} & {52.2} & {90.8 } & {54.8} & {51.3} & {57.6} & {40.8} & {58.4} & {40.5} & {57.7} \\
     {\checkmark} & {\checkmark} & {\checkmark} & {} & {\checkmark} & {} & {52.1} & {89.8} & {55.8} & {50.8} & {58.3} & {42} & {59.2} & {39.4} & {57.7} \\
     {\checkmark} & {\checkmark} & {\checkmark} & {} & {} & {\checkmark} & {57.3} & {90.8} & {63.8} & {56.4} & {65.7} & {45.6} & {63.3} & {47.3} & {61.3} \\
        
    \specialrule{0.1em}{1pt}{1pt}
    \end{tabular}}
    \end{table*}

\textbf{Effectiveness of components}. As shown in table~\ref{tab4}, the ”Sum” means element-wise summation for feature fusion, and the ”Cat” indicates that the features are fused by stacking them along the channel dimension. The training epochs are all 84. These results demonstrate the effectiveness of multi-view attentions and that different attentions are good at improving different segmentation metrics. It shows the complementarity of multi-view attention. Directly fusing them by either element-wise summation or stacking them leads to a decrease in $\rm mAP$ accuracy. The main reason is that straightforward feature fusion lacks the explicit fusion principle to preserve the typical view feature of weak objects. Our MHF achieves the best $\rm mAP$ and $\rm AP_s$ accuracies, which shows its effectiveness in representing weak objects using multi-view features.

\begin{figure*}[!htbp]
    \centering
    \includegraphics[width=6.0in]{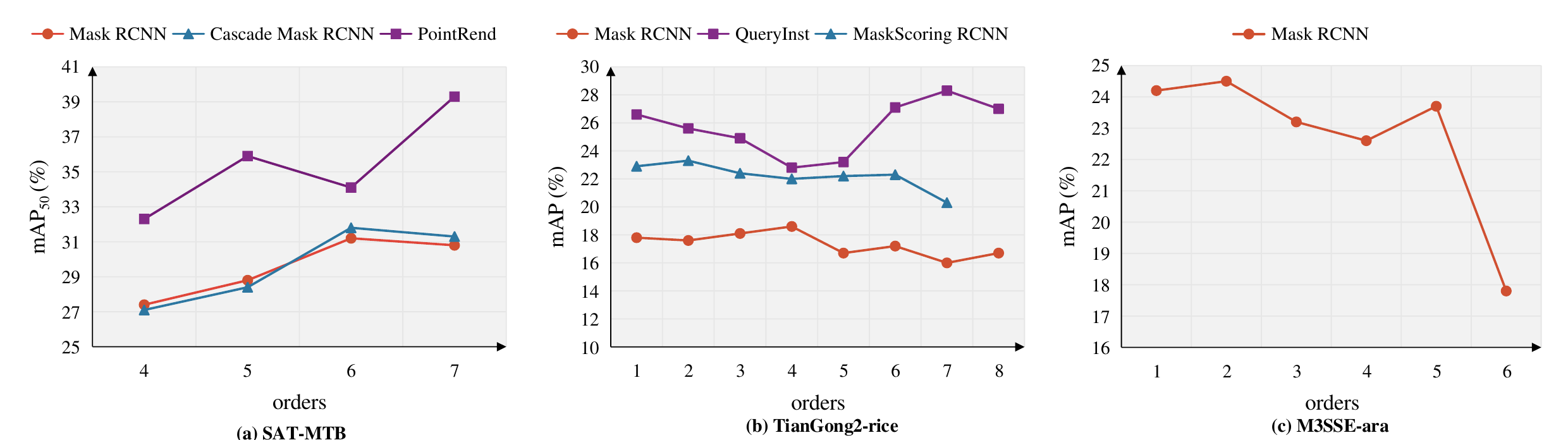}
    \caption{Object segmentation results of Mask RCNN equipped with our MHF with different order settings on the different dataset.}
    \label{fig4}
\end{figure*}

\textbf{Impacts of orders}. We also explore the impact of order settings for different methods on different datasets. The results are shown in Figure~\ref {fig4}. We report their segmentation results of $\rm mAP_{50}$ on the SAT-MTB dataset and mAP on the M3SSE-ara and TianGong2-rice datasets. We train different models for 12 epochs on the SAT-MTB and M3SSE-ara datasets and 120 epochs on the TianGong2-rice dataset. These results show that as the order values gradually increase, the overall accuracies increase first and then decrease. The order values of peak accuracies vary with different methods. The best order value of the same method is different on different datasets.  But the value of the best order usually lies in a limited range. We can dynamically adjust it to get better performance in weak object detection and segmentation according to different applications.

\begin{table*}[!htbp] 
    \footnotesize
    \renewcommand\arraystretch{0.95}
    \centering
    \caption{Object Segmentation results of Mask RCNN with different attentions on the M3SSE-ara dataset. 
    The "Sum" means element-wise summation operation for feature fusion.
    The "Cat" indicates that the features are fused by stacking them along the channel dimension.$f^s$, $f^c$, and $f^r$ are our spatial attention, channel attention, and frequency attention respectively.\label{tab5}}
    \setlength{\tabcolsep}{4.5mm}{
    \begin{tabular}{c|c|c|c|c}
                    
    \specialrule{0.1em}{1pt}{1pt}
    {\textbf{Method}} & {mAP} & {\textbf{Params(M)}} & {\textbf{FLOPs(G)}} & {\textbf{FPS}}  \\
    \hline
    {Baseline} & {51.6} & {43.76} & {389.46} & {14.3} \\
    \hline
    {SE~\cite{ref74}} & {53.2} & {43.82} & {389.5} & {13.6} \\
    {SG~\cite{ref113}} & {53.8} & {43.76} & {389.5} & {14.3} \\
    {SK~\cite{ref114}} & {54.2} & {65.96} & {1176.63} & {8.5} \\
    {CBAM~\cite{ref28}} & {51.7} & {43.79} & {389.54} & {12.6} \\
    {ECA~\cite{ref76}} & {53.6} & {43.76} & {389.5} & {14.2} \\
    {SF~\cite{ref115}} & {51.7} & {43.76} & {389.52} & {13.8} \\
    {ACmix~\cite{ref116}} & {51.4} & {44.63} & {421.64} & {5} \\
    {P-Self~\cite{ref117}} & {51.3} & {44.29} & {403.62} & {13.5} \\
    {CoT~\cite{ref118}} & {52.5} & {46.06} & {471.97} & {12} \\
    {TPT~\cite{ref30}} & {52.2} & {43.76} & {389.51} & {12.8} \\
    {PNT~\cite{ref119}} & {51.8} & {46.65} & {483.36} & {12.5} \\
    {Sum(SG,SK,ECA)} & {52.2} & {65.96} & {1176.71} & {8.3} \\
    {CAT(SG,SK,ECA)} & {51.5} & {66.74} & {1204.84} & {8.1} \\
    \hline
    {$f^s$} & {54.3} & {44.03} & {399.27} & {13.8} \\
    {$f^c$} & {54.7} & {43.77} & {389.83} & {13.6} \\
    {$f^r$} & {55.0} & {51.84} & {389.63} & {13.3} \\
    \hline
    {MHF(ours)} & {57.3} & {54.53} & {454.6} & {9.8} \\
               
    \specialrule{0.1em}{1pt}{1pt}
    \end{tabular}}
    \end{table*}

\textbf{Comparisons with Advanced Attentions}.
We compare our MHF with 11 latest and advanced attentions on the M3SSE-Ara dataset, using the Mask RCNN as the baseline. The backbone is ResNet50. The training epochs are 84. The insert location of different attentions are same as our MHF, which is after the feature extraction module. The attentions include SE~\cite{ref74}, SG~\cite{ref113}, SK~\cite{ref114}, CBAM~\cite{ref28}, ECA~\cite{ref76}, SF~\cite{ref115}, ACmix~\cite{ref116}, P-Self~\cite{ref117}, CoT~\cite{ref118}, TPT~\cite{ref30}, and PNT~\cite{ref119}. We report their segmentation results in Table~\ref {tab5}. We can see that not all attentions help improve the weak object segmentation accuracy on the M3SSE-ara dataset, such as ACmix and P-Self. And SG, SK, and ECA are top-3 attention on advancing the mAP accuracy of weak object segmentation. However, when we use the commonly used element-wise summation (Sum) and the channel stacking (Cat) to fuse them, the mAP accuracy decreases. Our MHF shows significant superiority in learning accurate and discriminative representations of weak objects. 

\section{Conclusion}
Weak objects are common in images and videos of space science and earth observation applications. Extracting accurate information about them through object detection and segmentation has significant value in promoting novel scientific discovery and advancing technological innovation. However, it is hard to learn proper representation from the limited and distorted appearance information of weak objects. 
Inspired by multi-view / multi-modality learning, we attempt to efficiently select information that is relevant and complementary to weak objects to enrich them from the multi-view features of images. We develop simple and effective spatial attention, channel attention, and frequency attention, treating their outputs as multi-view features. We propose a lightweight and plug-and-play method for multi-view features high-order fusion (MHF). It enhances the model's capacity to capture relevant and complementary information about weak objects by introducing high-order multi-view feature perception and task contribution sensing. Theoretically, other multi-view features or combinations of attention can be fused using our MHF in a higher-order manner. Without considering the cost of multi-view attentions, our MHF only costs less than 2M additional parameters. We perform extensive experiments on two space science datasets and a large-scale satellite video dataset, and apply our MHF to 21 advanced methods. Their detection and segmentation accuracies are significantly improved. These results clearly demonstrate the effectiveness and generality of our designs. In the future, we will further explore novel designs of multi-view features, feature selection, and feature fusion to advance the detection and segmentation accuracies of weak objects in space applications.

\section*{Acknowledgments}
Thanks for the scientific data from the National Basic Science Data Center ”Space Science and Application of China Manned Space Engineering DataBase”.



%

\bibliographystyle{IEEEtran}
\bibliography{refs}

@article{yangJ24,
  title={Video process detection for space electrostatic suspension material experiment in China’s Space Station},
  author={Yang, Jian and Liu, Kang and Zhao, Manqi and Li, Shengyang},
  journal={Engineering Applications of Artificial Intelligence},
  volume={131},
  pages={107804},
  year={2024},
  publisher={Elsevier}
}

@article{ref2,
  title={Multi-view Instance Attention Fusion Network for classification},
  author={Li, Jinxing and Zhou, Chuhao and Ji, Xiaoqiang and Li, Mu and Lu, Guangming and Xu, Yong and Zhang, David},
  journal={Information Fusion},
  volume={101},
  pages={101974},
  year={2024},
  publisher={Elsevier}
}

@inproceedings{ref3,
  title={Efficient Deep Palmprint Recognition via Distilled Hashing Coding},
  author={Shao, Huikai and Zhong, Dexing and Du, Xuefeng},
  booktitle={2019 IEEE/CVF Conference on Computer Vision and Pattern Recognition Workshops (CVPRW)},
  pages={714--723},
  year={2019},
  organization={IEEE Computer Society}
}

@article{ref4,
  title={Double-cohesion learning based multiview and discriminant palmprint recognition},
  author={Zhao, Shuping and Wu, Jigang and Fei, Lunke and Zhang, Bob and Zhao, Pengyang},
  journal={Information Fusion},
  volume={83},
  pages={96--109},
  year={2022},
  publisher={Elsevier}
}

@inproceedings{ref5,
  title={A probabilistic hierarchical model for multi-view and multi-feature classification},
  author={Li, Jinxing and Yong, Hongwei and Zhang, Bob and Li, Mu and Zhang, Lei and Zhang, David},
  booktitle={32nd AAAI Conference on Artificial Intelligence, AAAI 2018},
  pages={3498--3505},
  year={2018},
  organization={AAAI press}
}

@article{ref6,
  title={Shared autoencoder Gaussian process latent variable model for visual classification},
  author={Li, Jinxing and Zhang, Bob and Zhang, David},
  journal={IEEE transactions on neural networks and learning systems},
  volume={29},
  number={9},
  pages={4272--4286},
  year={2017},
  publisher={IEEE}
}

@article{ref7,
  title={Generative multi-view and multi-feature learning for classification},
  author={Li, Jinxing and Zhang, Bob and Lu, Guangming and Zhang, David},
  journal={Information Fusion},
  volume={45},
  pages={215--226},
  year={2019},
  publisher={Elsevier}
}

@article{ref8,
  title={Comprehensive multi-view representation learning},
  author={Zheng, Qinghai and Zhu, Jihua and Li, Zhongyu and Tian, Zhiqiang and Li, Chen},
  journal={Information Fusion},
  volume={89},
  pages={198--209},
  year={2023},
  publisher={Elsevier}
}

@inproceedings{ref9,
  title={Target-aware dual adversarial learning and a multi-scenario multi-modality benchmark to fuse infrared and visible for object detection},
  author={Liu, Jinyuan and Fan, Xin and Huang, Zhanbo and Wu, Guanyao and Liu, Risheng and Zhong, Wei and Luo, Zhongxuan},
  booktitle={Proceedings of the IEEE/CVF conference on computer vision and pattern recognition},
  pages={5802--5811},
  year={2022}
}

@article{ref10,
  title={Infrared and visible image fusion using discrete cosine transform and swarm intelligence for surveillance applications},
  author={Paramanandham, Nirmala and Rajendiran, Kishore},
  journal={Infrared Physics \& Technology},
  volume={88},
  pages={13--22},
  year={2018},
  publisher={Elsevier}
}

@article{ref11,
  title={Object classification using CNN-based fusion of vision and LIDAR in autonomous vehicle environment},
  author={Gao, Hongbo and Cheng, Bo and Wang, Jianqiang and Li, Keqiang and Zhao, Jianhui and Li, Deyi},
  journal={IEEE Transactions on Industrial Informatics},
  volume={14},
  number={9},
  pages={4224--4231},
  year={2018},
  publisher={IEEE}
}

@inproceedings{ref12,
  title={Multi-view super vector for action recognition},
  author={Cai, Zhuowei and Wang, Limin and Peng, Xiaojiang and Qiao, Yu},
  booktitle={Proceedings of the IEEE conference on Computer Vision and Pattern Recognition},
  pages={596--603},
  year={2014}
}

@inproceedings{ref13,
  title={A closed form solution to multi-view low-rank regression},
  author={Zheng, Shuai and Cai, Xiao and Ding, Chris and Nie, Feiping and Huang, Heng},
  booktitle={Proceedings of the Twenty-Ninth AAAI Conference on Artificial Intelligence},
  pages={1973--1979},
  year={2015}
}

@article{ref14,
  title={Visual attention methods in deep learning: An in-depth survey},
  author={Hassanin, Mohammed and Anwar, Saeed and Radwan, Ibrahim and Khan, Fahad Shahbaz and Mian, Ajmal},
  journal={Information Fusion},
  volume={108},
  pages={102417},
  year={2024},
  publisher={Elsevier}
}

@article{ref15,
  title={Satellite Video Multi-Label Scene Classification With Spatial and Temporal Feature Cooperative Encoding: A Benchmark Dataset and Method},
  author={Guo, Weilong and Li, Shengyang and Chen, Feixiang and Sun, Yuhan and Gu, Yanfeng},
  journal={IEEE Transactions on Image Processing},
  volume={33},
  pages={2238--2251},
  year={2024},
  publisher={IEEE}
}

@inproceedings{ref16,
  title={Rotation-Lossless Non-Local Attention and Fine-Grained Feature Enhancement Based Fine-Grained Oriented Object Detection in Remote Sensing Images},
  author={Guo, Weilong and Li, Xuan and Li, Shengyang},
  booktitle={2023 5th International Conference on Control and Robotics (ICCR)},
  pages={277--286},
  year={2023},
  organization={IEEE}
}

@article{ref17,
  title={Few-shot aircraft detection in satellite videos based on feature scale selection pyramid and proposal contrastive learning},
  author={Zhou, Zhuang and Li, Shengyang and Guo, Weilong and Gu, Yanfeng},
  journal={Remote Sensing},
  volume={14},
  number={18},
  pages={4581},
  year={2022},
  publisher={MDPI}
}

@article{ref18,
  title={Attention is all you need: utilizing attention in AI-enabled drug discovery},
  author={Zhang, Yang and Liu, Caiqi and Liu, Mujiexin and Liu, Tianyuan and Lin, Hao and Huang, Cheng-Bing and Ning, Lin},
  journal={Briefings in Bioinformatics},
  volume={25},
  number={1},
  pages={1--22},
  year={2024}
}

@inproceedings{ref19,
  title={Medical image segmentation via cascaded attention decoding},
  author={Rahman, Md Mostafijur and Marculescu, Radu},
  booktitle={Proceedings of the IEEE/CVF Winter Conference on Applications of Computer Vision},
  pages={6222--6231},
  year={2023}
}

@inproceedings{ref20,
  title={Medical Transformer: Gated Axial-Attention for Medical Image Segmentation},
  author={Valanarasu, Jeya Maria Jose and Oza, Poojan and Hacihaliloglu, Ilker and Patel, Vishal M},
  booktitle={International Conference on Medical Image Computing and Computer-Assisted Intervention},
  pages={36--46},
  year={2021}
}

@article{ref21,
  title={Crowd counting in smart city via lightweight ghost attention pyramid network},
  author={Guo, Xiangyu and Song, Kai and Gao, Mingliang and Zhai, Wenzhe and Li, Qilei and Jeon, Gwanggil},
  journal={Future Generation Computer Systems},
  volume={147},
  pages={328--338},
  year={2023},
  publisher={Elsevier}
}

@inproceedings{ref22,
  title={Uav-based multi-scale features fusion attention for fire detection in smart city ecosystems},
  author={Hussain, Tanveer and Dai, Hang and Gueaieb, Wail and Sicklinger, Marco and De Masi, Giulia},
  booktitle={2022 IEEE International Smart Cities Conference (ISC2)},
  pages={1--4},
  year={2022},
  organization={IEEE}
}

@article{ref23,
  title={Attention mechanisms in computer vision: A survey},
  author={Guo, Meng-Hao and Xu, Tian-Xing and Liu, Jiang-Jiang and Liu, Zheng-Ning and Jiang, Peng-Tao and Mu, Tai-Jiang and Zhang, Song-Hai and Martin, Ralph R and Cheng, Ming-Ming and Hu, Shi-Min},
  journal={Computational visual media},
  volume={8},
  number={3},
  pages={331--368},
  year={2022},
  publisher={Springer}
}

@article{ref24,
  title={MAE-BG: dual-stream boundary optimization for remote sensing image semantic segmentation},
  author={Yang, Ruiqi and Zheng, Chen and Wang, Leiguang and Zhao, Yili and Fu, Zhitao and Dai, Qinling},
  journal={Geocarto International},
  volume={38},
  number={1},
  pages={2190622},
  year={2023},
  publisher={Taylor \& Francis}
}

@article{ref25,
  title={Empowering Semantic Segmentation with Selective Frequency Enhancement and Attention Mechanism for Tampering Detection},
  author={Xu, Xu and Lv, Wenrui and Wang, Wei and Zhang, Yushu and Chen, Junxin},
  journal={IEEE Transactions on Artificial Intelligence},
  volume={1},
  number={01},
  pages={1--14},
  year={2023},
  publisher={IEEE Computer Society}
}

@inproceedings{ref26,
  title={MEGANet: Multi-Scale Edge-Guided Attention Network for Weak Boundary Polyp Segmentation},
  author={Bui, Nhat-Tan and Hoang, Dinh-Hieu and Nguyen, Quang-Thuc and Tran, Minh-Triet and Le, Ngan},
  booktitle={Proceedings of the IEEE/CVF Winter Conference on Applications of Computer Vision},
  pages={7985--7994},
  year={2024}
}

@inproceedings{ref27,
  title={Residual attention network for image classification},
  author={Wang, Fei and Jiang, Mengqing and Qian, Chen and Yang, Shuo and Li, Cheng and Zhang, Honggang and Wang, Xiaogang and Tang, Xiaoou},
  booktitle={Proceedings of the IEEE conference on computer vision and pattern recognition},
  pages={3156--3164},
  year={2017}
}

@inproceedings{ref28,
  title={Cbam: Convolutional block attention module},
  author={Woo, Sanghyun and Park, Jongchan and Lee, Joon-Young and Kweon, In So},
  booktitle={Proceedings of the European conference on computer vision (ECCV)},
  pages={3--19},
  year={2018}
}

@article{ref29,
  title={Recalibrating fully convolutional networks with spatial and channel “squeeze and excitation” blocks},
  author={Roy, Abhijit Guha and Navab, Nassir and Wachinger, Christian},
  journal={IEEE transactions on medical imaging},
  volume={38},
  number={2},
  pages={540--549},
  year={2018},
  publisher={IEEE}
}

@inproceedings{ref30,
  title={Rotate to attend: Convolutional triplet attention module},
  author={Misra, Diganta and Nalamada, Trikay and Arasanipalai, Ajay Uppili and Hou, Qibin},
  booktitle={Proceedings of the IEEE/CVF winter conference on applications of computer vision},
  pages={3139--3148},
  year={2021}
}

@inproceedings{ref31,
  title={Dual attention network for scene segmentation},
  author={Fu, Jun and Liu, Jing and Tian, Haijie and Li, Yong and Bao, Yongjun and Fang, Zhiwei and Lu, Hanqing},
  booktitle={Proceedings of the IEEE/CVF conference on computer vision and pattern recognition},
  pages={3146--3154},
  year={2019}
}

@inproceedings{ref32,
  title={Multi-interactive feature learning and a full-time multi-modality benchmark for image fusion and segmentation},
  author={Liu, Jinyuan and Liu, Zhu and Wu, Guanyao and Ma, Long and Liu, Risheng and Zhong, Wei and Luo, Zhongxuan and Fan, Xin},
  booktitle={Proceedings of the IEEE/CVF international conference on computer vision},
  pages={8115--8124},
  year={2023}
}

@article{ref33,
  title={Fusion of multispectral data through illumination-aware deep neural networks for pedestrian detection},
  author={Guan, Dayan and Cao, Yanpeng and Yang, Jiangxin and Cao, Yanlong and Yang, Michael Ying},
  journal={Information Fusion},
  volume={50},
  pages={148--157},
  year={2019},
  publisher={Elsevier}
}

@article{ref34,
  title={Revisiting feature fusion for RGB-T salient object detection},
  author={Zhang, Qiang and Xiao, Tonglin and Huang, Nianchang and Zhang, Dingwen and Han, Jungong},
  journal={IEEE Transactions on Circuits and Systems for Video Technology},
  volume={31},
  number={5},
  pages={1804--1818},
  year={2020},
  publisher={IEEE}
}

@ARTICLE{ref35,
  author={Liu, Risheng and Liu, Zhu and Liu, Jinyuan and Fan, Xin and Luo, Zhongxuan},
  journal={IEEE Transactions on Pattern Analysis and Machine Intelligence}, 
  title={A Task-guided, Implicitly-searched and Metainitialized Deep Model for Image Fusion}, 
  year={2024},
  volume={},
  number={},
  pages={1--16}
}

@article{ref36,
  title={Rtfnet: Rgb-thermal fusion network for semantic segmentation of urban scenes},
  author={Sun, Yuxiang and Zuo, Weixun and Liu, Ming},
  journal={IEEE Robotics and Automation Letters},
  volume={4},
  number={3},
  pages={2576--2583},
  year={2019},
  publisher={IEEE}
}

@article{ref37,
  title={Effective pan-sharpening by multiscale invertible neural network and heterogeneous task distilling},
  author={Zhou, Man and Huang, Jie and Fu, Xueyang and Zhao, Feng and Hong, Danfeng},
  journal={IEEE Transactions on Geoscience and Remote Sensing},
  volume={60},
  pages={1--14},
  year={2022},
  publisher={IEEE}
}

@inproceedings{ref38,
  title={MFNet: Towards real-time semantic segmentation for autonomous vehicles with multi-spectral scenes},
  author={Ha, Qishen and Watanabe, Kohei and Karasawa, Takumi and Ushiku, Yoshitaka and Harada, Tatsuya},
  booktitle={2017 IEEE/RSJ International Conference on Intelligent Robots and Systems (IROS)},
  pages={5108--5115},
  year={2017},
  organization={IEEE}
}

@inproceedings{ref39,
  title={Pst900: Rgb-thermal calibration, dataset and segmentation network},
  author={Shivakumar, Shreyas S and Rodrigues, Neil and Zhou, Alex and Miller, Ian D and Kumar, Vijay and Taylor, Camillo J},
  booktitle={2020 IEEE international conference on robotics and automation (ICRA)},
  pages={9441--9447},
  year={2020},
  organization={IEEE}
}

@article{ref40,
  title={CrossFuse: A novel cross attention mechanism based infrared and visible image fusion approach},
  author={Li, Hui and Wu, Xiao-Jun},
  journal={Information Fusion},
  volume={103},
  pages={102147},
  year={2024},
  publisher={Elsevier}
}

@article{ref41,
  title={MOFA: A novel dataset for Multi-modal Image Fusion Applications},
  author={Xiao, Kaihua and Kang, Xudong and Liu, Haibo and Duan, Puhong},
  journal={Information Fusion},
  volume={96},
  pages={144--155},
  year={2023},
  publisher={Elsevier}
}

@article{ref42,
  title={Deep learning on multi-view sequential data: a survey},
  author={Xie, Zhuyang and Yang, Yan and Zhang, Yiling and Wang, Jie and Du, Shengdong},
  journal={Artificial Intelligence Review},
  volume={56},
  number={7},
  pages={6661--6704},
  year={2023},
  publisher={Springer}
}

@inproceedings{ref43,
  title={High-order attention networks for medical image segmentation},
  author={Ding, Fei and Yang, Gang and Wu, Jun and Ding, Dayong and Xv, Jie and Cheng, Gangwei and Li, Xirong},
  booktitle={International Conference on Medical Image Computing and Computer-Assisted Intervention},
  pages={253--262},
  year={2020},
  organization={Springer}
}

@article{ref44,
  title={Hornet: Efficient high-order spatial interactions with recursive gated convolutions},
  author={Rao, Yongming and Zhao, Wenliang and Tang, Yansong and Zhou, Jie and Lim, Ser Nam and Lu, Jiwen},
  journal={Advances in Neural Information Processing Systems},
  volume={35},
  pages={10353--10366},
  year={2022}
}

@article{ref45,
  title={Focal Loss for Dense Object Detection},
  author={Lin, Tsung-Yi and Goyal, Priya and Girshick, Ross and He, Kaiming and Dollar, Piotr},
  journal={IEEE Transactions on Pattern Analysis \& Machine Intelligence},
  volume={42},
  number={02},
  pages={318--327},
  year={2020},
  publisher={IEEE Computer Society}
}

@INPROCEEDINGS{ref46,
  author={Tian, Zhi and Shen, Chunhua and Chen, Hao and He, Tong},
  booktitle={2019 IEEE/CVF International Conference on Computer Vision (ICCV)}, 
  title={FCOS: Fully Convolutional One-Stage Object Detection}, 
  year={2019},
  volume={},
  number={},
  pages={9626-9635}
}

@inproceedings{ref47,
  title={Reppoints: Point set representation for object detection},
  author={Yang, Ze and Liu, Shaohui and Hu, Han and Wang, Liwei and Lin, Stephen},
  booktitle={Proceedings of the IEEE/CVF international conference on computer vision},
  pages={9657--9666},
  year={2019}
}

@inproceedings{ref48,
  title={Yolact: Real-time instance segmentation},
  author={Bolya, Daniel and Zhou, Chong and Xiao, Fanyi and Lee, Yong Jae},
  booktitle={Proceedings of the IEEE/CVF international conference on computer vision},
  pages={9157--9166},
  year={2019}
}

@ARTICLE{ref49,
  author={Ren, Shaoqing and He, Kaiming and Girshick, Ross and Sun, Jian},
  journal={IEEE Transactions on Pattern Analysis and Machine Intelligence}, 
  title={Faster R-CNN: Towards Real-Time Object Detection with Region Proposal Networks}, 
  year={2017},
  volume={39},
  number={6},
  pages={1137-1149},
  }

@inproceedings{ref50,
  title={Cascade r-cnn: Delving into high quality object detection},
  author={Cai, Zhaowei and Vasconcelos, Nuno},
  booktitle={Proceedings of the IEEE conference on computer vision and pattern recognition},
  pages={6154--6162},
  year={2018}
}

@inproceedings{ref51,
  title={Mask r-cnn},
  author={He, Kaiming and Gkioxari, Georgia and Doll{\'a}r, Piotr and Girshick, Ross},
  booktitle={Proceedings of the IEEE international conference on computer vision},
  pages={2961--2969},
  year={2017}
}

@inproceedings{ref52,
  title={Grid r-cnn},
  author={Lu, Xin and Li, Buyu and Yue, Yuxin and Li, Quanquan and Yan, Junjie},
  booktitle={Proceedings of the IEEE/CVF conference on computer vision and pattern recognition},
  pages={7363--7372},
  year={2019}
}

@inproceedings{ref53,
  title={Libra r-cnn: Towards balanced learning for object detection},
  author={Pang, Jiangmiao and Chen, Kai and Shi, Jianping and Feng, Huajun and Ouyang, Wanli and Lin, Dahua},
  booktitle={Proceedings of the IEEE/CVF conference on computer vision and pattern recognition},
  pages={821--830},
  year={2019}
}

@inproceedings{ref54,
  title={Rethinking classification and localization for object detection},
  author={Wu, Yue and Chen, Yinpeng and Yuan, Lu and Liu, Zicheng and Wang, Lijuan and Li, Hongzhi and Fu, Yun},
  booktitle={Proceedings of the IEEE/CVF conference on computer vision and pattern recognition},
  pages={10186--10195},
  year={2020}
}

@inproceedings{ref55,
  title={Dynamic R-CNN: Towards high quality object detection via dynamic training},
  author={Zhang, Hongkai and Chang, Hong and Ma, Bingpeng and Wang, Naiyan and Chen, Xilin},
  booktitle={Computer Vision--ECCV 2020: 16th European Conference, Glasgow, UK, August 23--28, 2020, Proceedings, Part XV 16},
  pages={260--275},
  year={2020},
  organization={Springer}
}

@inproceedings{ref56,
  title={Sparse r-cnn: End-to-end object detection with learnable proposals},
  author={Sun, Peize and Zhang, Rufeng and Jiang, Yi and Kong, Tao and Xu, Chenfeng and Zhan, Wei and Tomizuka, Masayoshi and Li, Lei and Yuan, Zehuan and Wang, Changhu and others},
  booktitle={Proceedings of the IEEE/CVF conference on computer vision and pattern recognition},
  pages={14454--14463},
  year={2021}
}

@inproceedings{ref57,
  title={Hybrid task cascade for instance segmentation},
  author={Chen, Kai and Pang, Jiangmiao and Wang, Jiaqi and Xiong, Yu and Li, Xiaoxiao and Sun, Shuyang and Feng, Wansen and Liu, Ziwei and Shi, Jianping and Ouyang, Wanli and others},
  booktitle={Proceedings of the IEEE/CVF conference on computer vision and pattern recognition},
  pages={4974--4983},
  year={2019}
}

@inproceedings{ref58,
  title={Pointrend: Image segmentation as rendering},
  author={Kirillov, Alexander and Wu, Yuxin and He, Kaiming and Girshick, Ross},
  booktitle={Proceedings of the IEEE/CVF conference on computer vision and pattern recognition},
  pages={9799--9808},
  year={2020}
}

@inproceedings{ref59,
  title={End-to-end object detection with transformers},
  author={Carion, Nicolas and Massa, Francisco and Synnaeve, Gabriel and Usunier, Nicolas and Kirillov, Alexander and Zagoruyko, Sergey},
  booktitle={European conference on computer vision},
  pages={213--229},
  year={2020},
  organization={Springer}
}

@article{ref60,
  title={Dino: Detr with improved denoising anchor boxes for end-to-end object detection},
  author={Zhang, Hao and Li, Feng and Liu, Shilong and Zhang, Lei and Su, Hang and Zhu, Jun and Ni, Lionel M and Shum, Heung-Yeung},
  journal={arXiv preprint arXiv:2203.03605},
  year={2022}
}

@article{ref61,
  title={Object detection in 20 years: A survey},
  author={Zou, Zhengxia and Chen, Keyan and Shi, Zhenwei and Guo, Yuhong and Ye, Jieping},
  journal={Proceedings of the IEEE},
  volume={111},
  number={3},
  pages={257--276},
  year={2023},
  publisher={IEEE}
}

@inproceedings{ref62,
  title={Real-time single image and video super-resolution using an efficient sub-pixel convolutional neural network},
  author={Shi, Wenzhe and Caballero, Jose and Husz{\'a}r, Ferenc and Totz, Johannes and Aitken, Andrew P and Bishop, Rob and Rueckert, Daniel and Wang, Zehan},
  booktitle={Proceedings of the IEEE conference on computer vision and pattern recognition},
  pages={1874--1883},
  year={2016}
}

@article{ref63,
  title={Self-supervised feature augmentation for large image object detection},
  author={Pan, Xingjia and Tang, Fan and Dong, Weiming and Gu, Yang and Song, Zhichao and Meng, Yiping and Xu, Pengfei and Deussen, Oliver and Xu, Changsheng},
  journal={IEEE Transactions on Image Processing},
  volume={29},
  pages={6745--6758},
  year={2020},
  publisher={IEEE}
}

@article{ref64,
  title={Generative adversarial networks: An overview},
  author={Creswell, Antonia and White, Tom and Dumoulin, Vincent and Arulkumaran, Kai and Sengupta, Biswa and Bharath, Anil A},
  journal={IEEE signal processing magazine},
  volume={35},
  number={1},
  pages={53--65},
  year={2018},
  publisher={IEEE}
}

@article{ref65,
  title={Small-object detection in remote sensing images with end-to-end edge-enhanced GAN and object detector network},
  author={Rabbi, Jakaria and Ray, Nilanjan and Schubert, Matthias and Chowdhury, Subir and Chao, Dennis},
  journal={Remote Sensing},
  volume={12},
  number={9},
  pages={1432},
  year={2020},
  publisher={MDPI}
}

@article{ref66,
  title={Small object detection in remote sensing images with residual feature aggregation-based super-resolution and object detector network},
  author={Bashir, Syed Muhammad Arsalan and Wang, Yi},
  journal={Remote Sensing},
  volume={13},
  number={9},
  pages={1854},
  year={2021},
  publisher={MDPI}
}

@inproceedings{ref67,
  title={Sod-mtgan: Small object detection via multi-task generative adversarial network},
  author={Bai, Yancheng and Zhang, Yongqiang and Ding, Mingli and Ghanem, Bernard},
  booktitle={Proceedings of the European conference on computer vision (ECCV)},
  pages={206--221},
  year={2018}
}

@article{ref68,
  title={Object detection in optical remote sensing images: A survey and a new benchmark},
  author={Li, Ke and Wan, Gang and Cheng, Gong and Meng, Liqiu and Han, Junwei},
  journal={ISPRS journal of photogrammetry and remote sensing},
  volume={159},
  pages={296--307},
  year={2020},
  publisher={Elsevier}
}

@article{ref69,
  title={Ship rotated bounding box space for ship extraction from high-resolution optical satellite images with complex backgrounds},
  author={Liu, Zikun and Wang, Hongzhen and Weng, Lubin and Yang, Yiping},
  journal={IEEE geoscience and remote sensing letters},
  volume={13},
  number={8},
  pages={1074--1078},
  year={2016},
  publisher={IEEE}
}

@article{ref70,
  title={Feature Enhancement Network for Object Detection in Optical Remote Sensing Images},
  author={Cheng, Gong and Lang, Chunbo and Wu, Maoxiong and Xie, Xingxing and Yao, Xiwen and Han, Junwei},
  journal={Journal of Remote Sensing},
  volume={2021},
  pages={9805389},
  year={2021}
}

@inproceedings{ref71,
  title={Cascade detector with feature fusion for arbitrary-oriented objects in remote sensing images},
  author={Hou, Liping and Lu, Ke and Xue, Jian and Hao, Li},
  booktitle={2020 IEEE International Conference on Multimedia and Expo (ICME)},
  pages={1--6},
  year={2020},
  organization={IEEE}
}

@article{ref72,
  title={Towards Large-Scale Small Object Detection: Survey and Benchmarks},
  author={Cheng, Gong and Yuan, Xiang and Yao, Xiwen and Yan, Kebing and Zeng, Qinghua and Xie, Xingxing and Han, Junwei},
  journal={IEEE Transactions on Pattern Analysis \& Machine Intelligence},
  volume={45},
  number={11},
  pages={13467--13488},
  year={2023},
  publisher={IEEE Computer Society}
}

@inproceedings{ref73,
  title={Sca-cnn: Spatial and channel-wise attention in convolutional networks for image captioning},
  author={Chen, Long and Zhang, Hanwang and Xiao, Jun and Nie, Liqiang and Shao, Jian and Liu, Wei and Chua, Tat-Seng},
  booktitle={Proceedings of the IEEE conference on computer vision and pattern recognition},
  pages={5659--5667},
  year={2017}
}

@inproceedings{ref74,
  title={Squeeze-and-excitation networks},
  author={Hu, Jie and Shen, Li and Sun, Gang},
  booktitle={Proceedings of the IEEE conference on computer vision and pattern recognition},
  pages={7132--7141},
  year={2018}
}

@inproceedings{ref75,
  title={Global second-order pooling convolutional networks},
  author={Gao, Zilin and Xie, Jiangtao and Wang, Qilong and Li, Peihua},
  booktitle={Proceedings of the IEEE/CVF Conference on computer vision and pattern recognition},
  pages={3024--3033},
  year={2019}
}

@inproceedings{ref76,
  title={ECA-Net: Efficient channel attention for deep convolutional neural networks},
  author={Wang, Qilong and Wu, Banggu and Zhu, Pengfei and Li, Peihua and Zuo, Wangmeng and Hu, Qinghua},
  booktitle={Proceedings of the IEEE/CVF conference on computer vision and pattern recognition},
  pages={11534--11542},
  year={2020}
}

@inproceedings{ref77,
  title={Srm: A style-based recalibration module for convolutional neural networks},
  author={Lee, HyunJae and Kim, Hyo-Eun and Nam, Hyeonseob},
  booktitle={Proceedings of the IEEE/CVF International conference on computer vision},
  pages={1854--1862},
  year={2019}
}

@inproceedings{ref78,
  title={Recurrent models of visual attention},
  author={Mnih, Volodymyr and Heess, Nicolas and Graves, Alex and Kavukcuoglu, Koray},
  booktitle={Proceedings of the 27th International Conference on Neural Information Processing Systems-Volume 2},
  pages={2204--2212},
  year={2014}
}

@inproceedings{ref79,
  title={Spatial transformer networks},
  author={Jaderberg, Max and Simonyan, Karen and Zisserman, Andrew and Kavukcuoglu, Koray},
  booktitle={Proceedings of the 28th International Conference on Neural Information Processing Systems-Volume 2},
  pages={2017--2025},
  year={2015}
}

@inproceedings{ref80,
  title={Gather-excite: exploiting feature context in convolutional neural networks},
  author={Hu, Jie and Shen, Li and Albanie, Samuel and Sun, Gang and Vedaldi, Andrea},
  booktitle={Proceedings of the 32nd International Conference on Neural Information Processing Systems},
  pages={9423--9433},
  year={2018}
}

@inproceedings{ref81,
  title={Non-local neural networks},
  author={Wang, Xiaolong and Girshick, Ross and Gupta, Abhinav and He, Kaiming},
  booktitle={Proceedings of the IEEE conference on computer vision and pattern recognition},
  pages={7794--7803},
  year={2018}
}

@inproceedings{ref82,
  title={Deep residual learning for image recognition},
  author={He, Kaiming and Zhang, Xiangyu and Ren, Shaoqing and Sun, Jian},
  booktitle={Proceedings of the IEEE conference on computer vision and pattern recognition},
  pages={770--778},
  year={2016}
}

@inproceedings{ref83,
  title={Yolov3: An incremental improvement},
  author={Farhadi, Ali and Redmon, Joseph},
  booktitle={Computer vision and pattern recognition},
  volume={1804},
  pages={1--6},
  year={2018},
  organization={Springer Berlin/Heidelberg, Germany}
}

@inproceedings{ref84,
  title={Feature pyramid networks for object detection},
  author={Lin, Tsung-Yi and Doll{\'a}r, Piotr and Girshick, Ross and He, Kaiming and Hariharan, Bharath and Belongie, Serge},
  booktitle={Proceedings of the IEEE conference on computer vision and pattern recognition},
  pages={2117--2125},
  year={2017}
}

@inproceedings{ref85,
  title={You only look once: Unified, real-time object detection},
  author={Redmon, Joseph and Divvala, Santosh and Girshick, Ross and Farhadi, Ali},
  booktitle={Proceedings of the IEEE conference on computer vision and pattern recognition},
  pages={779--788},
  year={2016}
}

@article{ref86,
  title={A Multitask Benchmark Dataset for Satellite Video: Object Detection, Tracking, and Segmentation},
  author={Li, Shengyang and Zhou, Zhuang and Zhao, Manqi and Yang, Jian and Guo, Weilong and Lv, Yixuan and Kou, Longxuan and Wang, Han and Gu, Yanfeng},
  journal={IEEE Transactions on Geoscience and Remote Sensing},
  volume={61},
  pages={3278075},
  year={2023}
}

@inproceedings{ref87,
  title={Microsoft coco: Common objects in context},
  author={Lin, Tsung-Yi and Maire, Michael and Belongie, Serge and Hays, James and Perona, Pietro and Ramanan, Deva and Doll{\'a}r, Piotr and Zitnick, C Lawrence},
  booktitle={Computer Vision--ECCV 2014: 13th European Conference, Zurich, Switzerland, September 6-12, 2014, Proceedings, Part V 13},
  pages={740--755},
  year={2014},
  organization={Springer}
}

@inproceedings{ref88,
  title={Cornernet: Detecting objects as paired keypoints},
  author={Law, Hei and Deng, Jia},
  booktitle={Proceedings of the European conference on computer vision (ECCV)},
  pages={734--750},
  year={2018}
}

@article{ref89,
  title={Objects as points},
  author={Zhou, Xingyi and Wang, Dequan and Kr{\"a}henb{\"u}hl, Philipp},
  journal={arXiv preprint arXiv:1904.07850},
  year={2019}
}

@inproceedings{ref90,
  title={An empirical study of spatial attention mechanisms in deep networks},
  author={Zhu, Xizhou and Cheng, Dazhi and Zhang, Zheng and Lin, Stephen and Dai, Jifeng},
  booktitle={Proceedings of the IEEE/CVF international conference on computer vision},
  pages={6688--6697},
  year={2019}
}

@inproceedings{ref91,
  title={Centripetalnet: Pursuing high-quality keypoint pairs for object detection},
  author={Dong, Zhiwei and Li, Guoxuan and Liao, Yue and Wang, Fei and Ren, Pengju and Qian, Chen},
  booktitle={Proceedings of the IEEE/CVF conference on computer vision and pattern recognition},
  pages={10519--10528},
  year={2020}
}

@inproceedings{ref92,
  title={Feature selective anchor-free module for single-shot object detection},
  author={Zhu, Chenchen and He, Yihui and Savvides, Marios},
  booktitle={Proceedings of the IEEE/CVF conference on computer vision and pattern recognition},
  pages={840--849},
  year={2019}
}

@inproceedings{ref93,
  title={Bridging the gap between anchor-based and anchor-free detection via adaptive training sample selection},
  author={Zhang, Shifeng and Chi, Cheng and Yao, Yongqiang and Lei, Zhen and Li, Stan Z},
  booktitle={Proceedings of the IEEE/CVF conference on computer vision and pattern recognition},
  pages={9759--9768},
  year={2020}
}

@article{ref94,
  title={Foveabox: Beyound anchor-based object detection},
  author={Kong, Tao and Sun, Fuchun and Liu, Huaping and Jiang, Yuning and Li, Lei and Shi, Jianbo},
  journal={IEEE Transactions on Image Processing},
  volume={29},
  pages={7389--7398},
  year={2020},
  publisher={IEEE}
}

@inproceedings{ref95,
  title={Scnet: Training inference sample consistency for instance segmentation},
  author={Vu, Thang and Kang, Haeyong and Yoo, Chang D},
  booktitle={Proceedings of the AAAI Conference on Artificial Intelligence},
  volume={35},
  number={3},
  pages={2701--2709},
  year={2021}
}

@inproceedings{ref96,
  title={Mask scoring r-cnn},
  author={Huang, Zhaojin and Huang, Lichao and Gong, Yongchao and Huang, Chang and Wang, Xinggang},
  booktitle={Proceedings of the IEEE/CVF conference on computer vision and pattern recognition},
  pages={6409--6418},
  year={2019}
}

@inproceedings{ref97,
  title={Instances as queries},
  author={Fang, Yuxin and Yang, Shusheng and Wang, Xinggang and Li, Yu and Fang, Chen and Shan, Ying and Feng, Bin and Liu, Wenyu},
  booktitle={Proceedings of the IEEE/CVF international conference on computer vision},
  pages={6910--6919},
  year={2021}
}

@inproceedings{ref98,
  title={Sparse instance activation for real-time instance segmentation},
  author={Cheng, Tianheng and Wang, Xinggang and Chen, Shaoyu and Zhang, Wenqiang and Zhang, Qian and Huang, Chang and Zhang, Zhaoxiang and Liu, Wenyu},
  booktitle={Proceedings of the IEEE/CVF Conference on Computer Vision and Pattern Recognition},
  pages={4433--4442},
  year={2022}
}

@article{ref99,
  title={Generalized focal loss: Learning qualified and distributed bounding boxes for dense object detection},
  author={Li, Xiang and Wang, Wenhai and Wu, Lijun and Chen, Shuo and Hu, Xiaolin and Li, Jun and Tang, Jinhui and Yang, Jian},
  journal={Advances in Neural Information Processing Systems},
  volume={33},
  pages={21002--21012},
  year={2020}
}

@inproceedings{ref100,
  title={Dynamic head: Unifying object detection heads with attentions},
  author={Dai, Xiyang and Chen, Yinpeng and Xiao, Bin and Chen, Dongdong and Liu, Mengchen and Yuan, Lu and Zhang, Lei},
  booktitle={Proceedings of the IEEE/CVF conference on computer vision and pattern recognition},
  pages={7373--7382},
  year={2021}
}

@article{ref101,
  title={Dab-detr: Dynamic anchor boxes are better queries for detr},
  author={Liu, Shilong and Li, Feng and Zhang, Hao and Yang, Xiao and Qi, Xianbiao and Su, Hang and Zhu, Jun and Zhang, Lei},
  journal={arXiv preprint arXiv:2201.12329},
  year={2022}
}

@inproceedings{ref102,
  title={Detectors: Detecting objects with recursive feature pyramid and switchable atrous convolution},
  author={Qiao, Siyuan and Chen, Liang-Chieh and Yuille, Alan},
  booktitle={Proceedings of the IEEE/CVF conference on computer vision and pattern recognition},
  pages={10213--10224},
  year={2021}
}

@inproceedings{ref103,
  title={Flow-guided feature aggregation for video object detection},
  author={Zhu, Xizhou and Wang, Yujie and Dai, Jifeng and Yuan, Lu and Wei, Yichen},
  booktitle={Proceedings of the IEEE international conference on computer vision},
  pages={408--417},
  year={2017}
}

@inproceedings{ref104,
  title={Deep feature flow for video recognition},
  author={Zhu, Xizhou and Xiong, Yuwen and Dai, Jifeng and Yuan, Lu and Wei, Yichen},
  booktitle={Proceedings of the IEEE conference on computer vision and pattern recognition},
  pages={2349--2358},
  year={2017}
}

@inproceedings{ref105,
  title={Sequence level semantics aggregation for video object detection},
  author={Wu, Haiping and Chen, Yuntao and Wang, Naiyan and Zhang, Zhaoxiang},
  booktitle={Proceedings of the IEEE/CVF international conference on computer vision},
  pages={9217--9225},
  year={2019}
}

@inproceedings{ref106,
  title={Temporal ROI align for video object recognition},
  author={Gong, Tao and Chen, Kai and Wang, Xinjiang and Chu, Qi and Zhu, Feng and Lin, Dahua and Yu, Nenghai and Feng, Huamin},
  booktitle={Proceedings of the AAAI Conference on Artificial Intelligence},
  volume={35},
  number={2},
  pages={1442--1450},
  year={2021}
}

@inproceedings{ref107,
  title={Scale-aware trident networks for object detection},
  author={Li, Yanghao and Chen, Yuntao and Wang, Naiyan and Zhang, Zhaoxiang},
  booktitle={Proceedings of the IEEE/CVF international conference on computer vision},
  pages={6054--6063},
  year={2019}
}

@inproceedings{ref108,
  title={A novel region of interest extraction layer for instance segmentation},
  author={Rossi, Leonardo and Karimi, Akbar and Prati, Andrea},
  booktitle={2020 25th international conference on pattern recognition (ICPR)},
  pages={2203--2209},
  year={2021},
  organization={IEEE}
}

@inproceedings{ref109,
  title={Seqformer: Sequential transformer for video instance segmentation},
  author={Wu, Junfeng and Jiang, Yi and Bai, Song and Zhang, Wenqing and Bai, Xiang},
  booktitle={European Conference on Computer Vision},
  pages={553--569},
  year={2022},
  organization={Springer}
}

@inproceedings{ref110,
  title={In defense of online models for video instance segmentation},
  author={Wu, Junfeng and Liu, Qihao and Jiang, Yi and Bai, Song and Yuille, Alan and Bai, Xiang},
  booktitle={European Conference on Computer Vision},
  pages={588--605},
  year={2022},
  organization={Springer}
}

@inproceedings{ref111,
  title={Mask transfiner for high-quality instance segmentation},
  author={Ke, Lei and Danelljan, Martin and Li, Xia and Tai, Yu-Wing and Tang, Chi-Keung and Yu, Fisher},
  booktitle={Proceedings of the IEEE/CVF Conference on Computer Vision and Pattern Recognition},
  pages={4412--4421},
  year={2022}
}

@inproceedings{ref112,
  title={End-to-end referring video object segmentation with multimodal transformers},
  author={Botach, Adam and Zheltonozhskii, Evgenii and Baskin, Chaim},
  booktitle={Proceedings of the IEEE/CVF Conference on Computer Vision and Pattern Recognition},
  pages={4985--4995},
  year={2022}
}

@inproceedings{ref113,
  title={Spatial group-wise enhance: Enhancing semantic feature learning in cnn},
  author={Li, Yuxuan and Li, Xiang and Yang, Jian},
  booktitle={Proceedings of the Asian Conference on Computer Vision},
  pages={687--702},
  year={2022}
}

@inproceedings{ref114,
  title={Selective kernel networks},
  author={Li, Xiang and Wang, Wenhai and Hu, Xiaolin and Yang, Jian},
  booktitle={Proceedings of the IEEE/CVF conference on computer vision and pattern recognition},
  pages={510--519},
  year={2019}
}

@inproceedings{ref115,
  title={Sa-net: Shuffle attention for deep convolutional neural networks},
  author={Zhang, Qing-Long and Yang, Yu-Bin},
  booktitle={ICASSP 2021-2021 IEEE International Conference on Acoustics, Speech and Signal Processing (ICASSP)},
  pages={2235--2239},
  year={2021},
  organization={IEEE}
}

@inproceedings{ref116,
  title={On the integration of self-attention and convolution},
  author={Pan, Xuran and Ge, Chunjiang and Lu, Rui and Song, Shiji and Chen, Guanfu and Huang, Zeyi and Huang, Gao},
  booktitle={Proceedings of the IEEE/CVF conference on computer vision and pattern recognition},
  pages={815--825},
  year={2022}
}

@article{ref117,
  title={Polarized self-attention: Towards high-quality pixel-wise mapping},
  author={Liu, Huajun and Liu, Fuqiang and Fan, Xinyi and Huang, Dong},
  journal={Neurocomputing},
  volume={506},
  pages={158--167},
  year={2022},
  publisher={Elsevier}
}

@article{ref118,
  title={Contextual transformer networks for visual recognition},
  author={Li, Yehao and Yao, Ting and Pan, Yingwei and Mei, Tao},
  journal={IEEE Transactions on Pattern Analysis and Machine Intelligence},
  volume={45},
  number={2},
  pages={1489--1500},
  year={2022},
  publisher={IEEE}
}

@article{ref119,
  title={Non-deep networks},
  author={Goyal, Ankit and Bochkovskiy, Alexey and Deng, Jia and Koltun, Vladlen},
  journal={Advances in neural information processing systems},
  volume={35},
  pages={6789--6801},
  year={2022}
}

\vspace{11pt}

\begin{IEEEbiography}[{\includegraphics[width=1in,height=1.25in,clip,keepaspectratio]{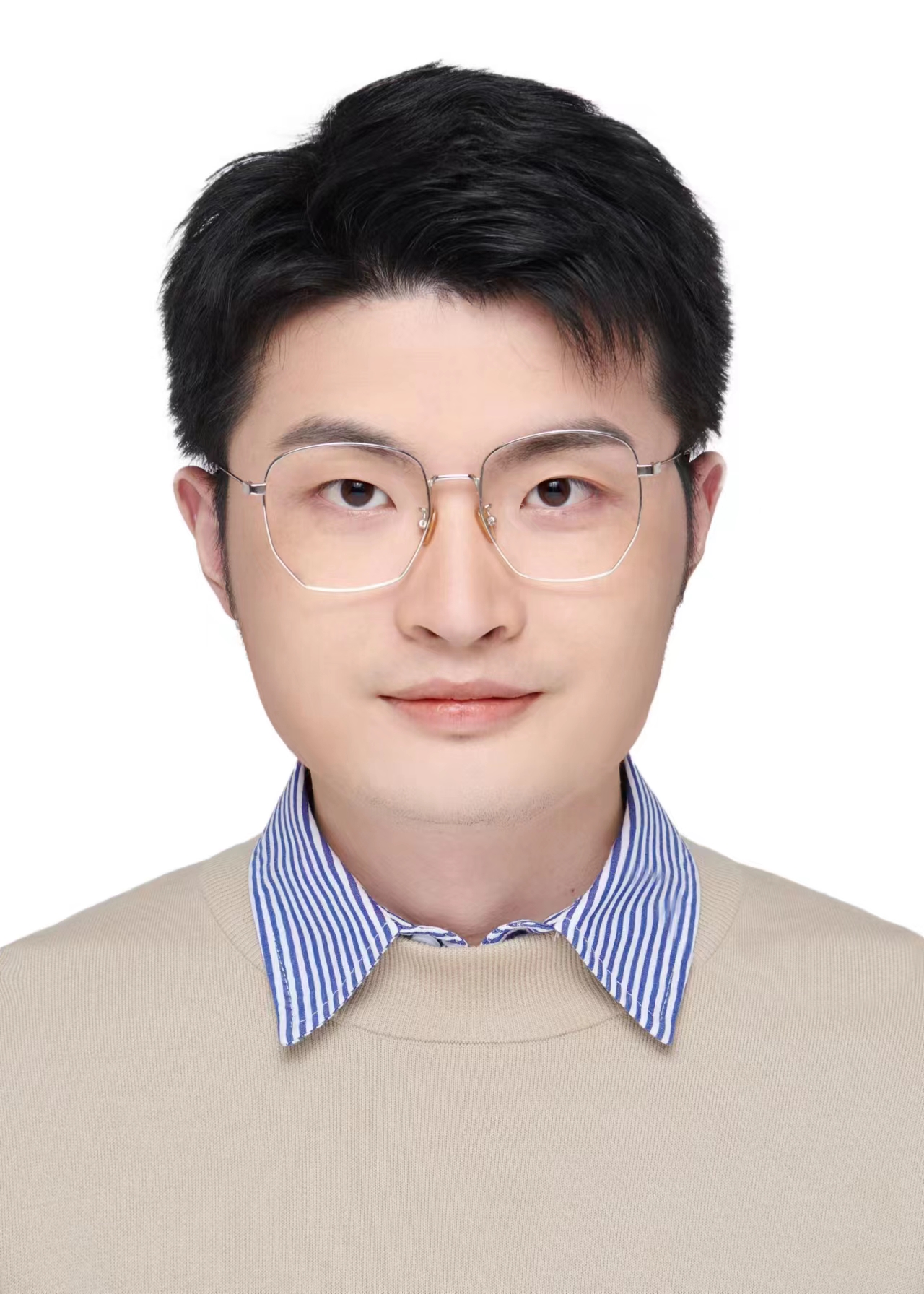}}]{Weilong Guo}
  earned his B.Eng. degree in software engineering from Jilin University, Jilin, China, in 2018, and subsequently, 
  his M.Eng. degree in computer applied technology from the School of Artificial Intelligence, University of Chinese Academy of Sciences, 
  Beijing, China, in 2022.

  Currently serving as an Engineer at the Technology and Engineering Center for Space Utilization, Chinese Academy of Sciences, Beijing, China, 
  he is deeply engaged in research focused on the intelligent analysis and understanding of images and videos.
\end{IEEEbiography}

\begin{IEEEbiography}[{\includegraphics[width=1in,height=1.25in,clip,keepaspectratio]{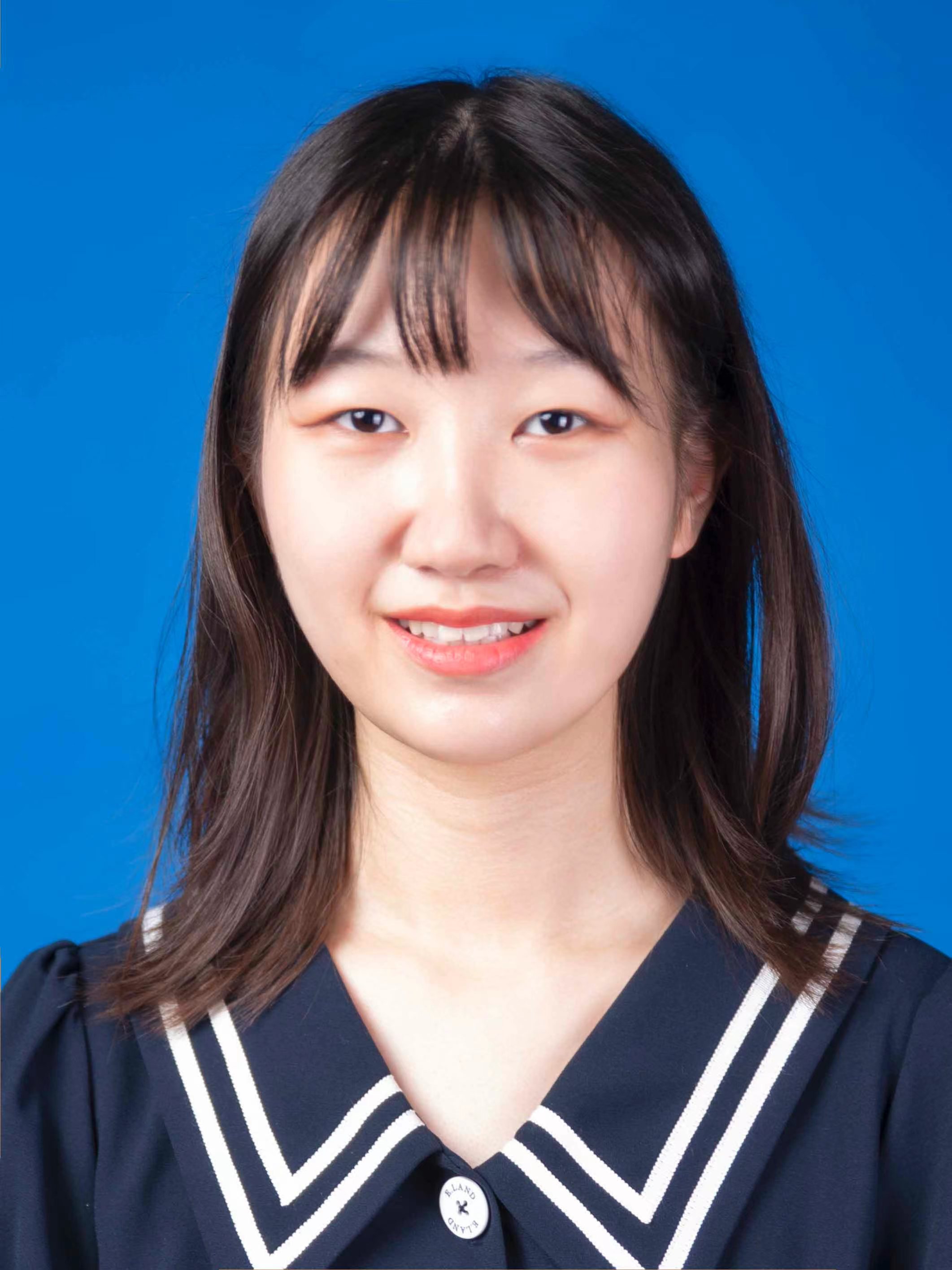}}]{Yuhan Sun}
  received the B.S. degree in Automation and Computer Science and Technology from Xi'an Jiaotong University, Xi'an, Shannxi, China, in 2021. 
  
She is currently pursing the Ph.D. degree in Computer-applied Technology with the Technology and Engineering Center for Space Utilization, Chinese Academy of Sciences (CAS), Beijing. Her research interests include satellite video and conventional video analysis, with a focus on object detection and segmentation.
\end{IEEEbiography}

\begin{IEEEbiography}[{\includegraphics[width=1in,height=1.25in,clip,keepaspectratio]{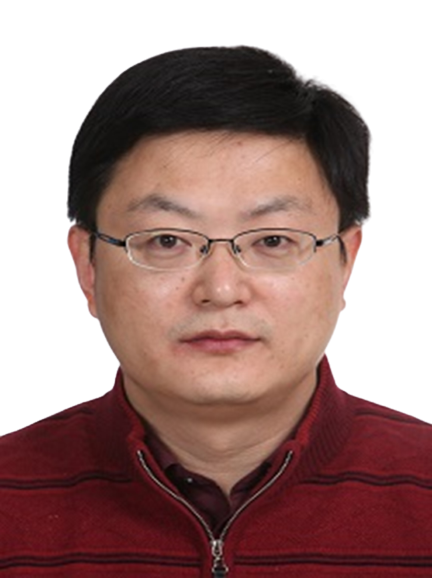}}]{Shengyang Li}
received the Ph.D. degree from the Institute of Remote Sensing Applications, Chinese Academy of Sciences, Beijing, China, in 2006.

He is currently a Professor with the Technology and Engineering Center for Space Utilization, Chinese Academy of Sciences. His research activities are machine learning in remote sensing image interpretation, deep learning in satellite videos processing and analysis, intelligent image processing, analysis and understanding for space utilization, and space  scientific big data modeling and analysis.
\end{IEEEbiography}

\vspace{11pt}

\vfill

\end{document}